\documentclass[preprint,12pt]{elsarticle}

\usepackage[ font=footnotesize,labelfont=bf,labelsep=none]{caption}
\usepackage{amssymb}
\biboptions{sort&compress}
\usepackage{graphicx}
\usepackage[tbtags]{amsmath}
\usepackage{array}
\usepackage{booktabs}
\usepackage{amsmath}
\usepackage{multirow}
\usepackage{color}
\usepackage{hyperref}
\hypersetup{colorlinks=true}
\usepackage{enumitem}
\usepackage{makecell}
\usepackage{float}
\usepackage[table]{xcolor}
\usepackage{colortbl}
\usepackage{subfigure}
\usepackage{bm}
\usepackage{balance}

\journal{Infrared Physics \& Technology}

\begin{document}

\begin{frontmatter}




\title{GAN-HA: A generative adversarial network with a novel heterogeneous dual-discriminator network and a new attention-based fusion strategy for infrared and visible image fusion}






\author{Guosheng Lu}
\author{Zile Fang}
\author{Jiaju Tian}
\author{Haowen Huang}
\author{Yuelong Xu}
\author{Zhuolin Han}
\author{Yaoming Kang}
\author{Can Feng}
\author{Zhigang Zhao\corref{mycorrespondingauthor}}
\cortext[mycorrespondingauthor]{Corresponding author}
\ead{zhaozhigang@sztu.edu.cn}

\address{College of New Materials and New Energies, Shenzhen Technology University, Shenzhen, 518118, China}


\begin{abstract}
Infrared and visible image fusion (IVIF) aims to preserve thermal radiation information from infrared images while integrating texture details from visible images. Thermal radiation information is mainly expressed through image intensities, while texture details are typically expressed through image gradients. However, existing dual-discriminator generative adversarial networks (GANs) often rely on two structurally identical discriminators for learning, which do not fully account for the distinct learning needs of infrared and visible image information. To this end, this paper proposes a novel GAN with a heterogeneous dual-discriminator network and an attention-based fusion strategy (GAN-HA). Specifically, recognizing the intrinsic differences between infrared and visible images, we propose, for the first time, a novel heterogeneous dual-discriminator network to simultaneously capture thermal radiation information and texture details. The two discriminators in this network are structurally different, including a salient discriminator for infrared images and a detailed discriminator for visible images. They are able to learn rich image intensity information and image gradient information, respectively. In addition, a new attention-based fusion strategy is designed in the generator to appropriately emphasize the learned information from different source images, thereby improving the information representation ability of the fusion result. In this way, the fused images generated by GAN-HA can more effectively maintain both the salience of thermal targets and the sharpness of textures. Extensive experiments on various public datasets demonstrate the superiority of GAN-HA over other state-of-the-art (SOTA) algorithms while showcasing its higher potential for practical applications. 
\end{abstract}

\end{frontmatter}

\section{Introduction}
\label{section:A}
\begin{figure}
\setlength{\abovecaptionskip}{0.1cm}
\setlength{\belowcaptionskip}{-0.5cm}
  \centering
  \includegraphics[width=\linewidth]{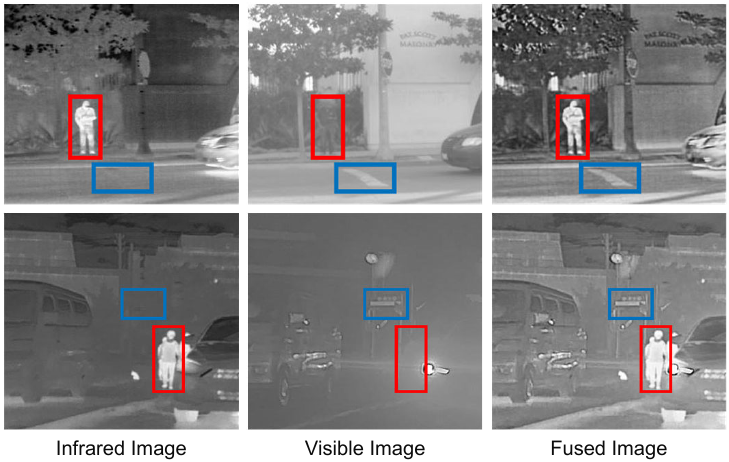}
  \caption{Example of the effect of IVIF.}
  \label{Fig.1}
\end{figure}

\vspace{-0.2cm}
Infrared and visible image fusion (IVIF) provides superior complementarity to the images produced by both sensors \cite{pan2024residual, xiong2024segfusion, chi2024lmdfusion, zhang2021gan}. Infrared sensors capture thermal radiation information emitted by objects, effectively highlighting significant targets even in difficult conditions such as fog, poor illumination, and other bad weather \cite{ma2020infrared, jin2017survey}. However, these images typically exhibit lower spatial resolution \cite{wang2024raw}. Visible sensors capture reflected light information, yielding higher spatial resolution that can effectively characterize details in complex scenes \cite{yang2024review}. Nevertheless, visible images are easily influenced by severe conditions \cite{jin2017survey, ma2019infrared}. By fusing these two types of images, their respective shortcomings can be effectively compensated so that fusion results can present comprehensive and important information in harsh environments or complex scenes. Fig. \ref{Fig.1} illustrates the benefits of IVIF, underscoring its high potential for practical engineering applications \cite{liang2024ifici, zhang2021gan}.

In recent decades, various approaches to image fusion have emerged, typically divided into two categories: model-based methods and deep learning-based methods.

Model-based methods usually fall into four categories: multi-scale transform methods (MST) \cite{du2016union, he2019image, lewis2007pixel, nencini2007remote}, sparse representation (SR) \cite{zhang2018robust, wang2014fusion, liu2016image, he2023camouflaged}, subspace methods \cite{mishra2014mri, cvejic2007region}, and hybrid techniques \cite{liu2015general, he2024weakly, he2023strategic}. Despite their ability to yield satisfactory results, these methods suffer from inherent limitations. Specifically, manually designed feature extraction operators and fusion rules often fail to capture image features comprehensively. 

Conversely, deep learning has gained prominence in image fusion tasks due to the robust data representation capabilities of neural networks. The primary deep learning-based fusion methods include convolutional neural network (CNN) methods \cite{li2021infrared, tang2022mdedfusion, xing2024cfnet, xiong2024segfusion, xiao2023concealed, tang2023source, ju2022ivf, liu2017multi, li2019infrared, zhang2020ifcnn}, auto-encoder (AE) methods \cite{li2018densefuse, tang2023consistency, xu2022multi, li2020nestfuse, wang2021unfusion}, and generative adversarial network (GAN) methods \cite{zhang2021gan, ma2019fusiongan, ma2020ddcgan, ma2020ganmcc, li2020attentionfgan, li2020multigrained, liu2022target, deng2022pcgan, he2023hqg, zhou2021semantic, xiao2024survey, he2024diffusion, fang2024real}. Due to the absence of ground truth or standard fused images in image fusion tasks, GAN methods can perform better by using source images to guide model learning. In particular, dual-discriminator GAN methods, which use two independent discriminator structures to learn the features of infrared and visible images separately, are particularly well-suited for multi-source fusion tasks. These methods can generate high-quality fused images that contain both infrared information and visible visual details. Many dual-discriminator GAN models have been proposed \cite{li2020multigrained, deng2022pcgan, he2023hqg, zhang2021gan, li2020attentionfgan, liu2022target, zhou2021semantic}, all showing remarkable fusion performance.

Nevertheless, several aspects still have drawbacks. Firstly, there are inherent differences between the information representation of infrared and visible images. Infrared thermal radiation is characterized by image intensities, while texture details are characterized by edges and gradients. These two types of information are significantly different. Existing dual-discriminator GAN methods overlook this issue and use two structurally identical discriminators to encourage model learning. This inevitably makes it difficult for the model to sufficiently learn different types of information simultaneously, causing the fusion result to sacrifice some thermal radiation information or texture details. Secondly, current methods mainly focus on information extraction, while less research considers the fusion of extracted information \cite{zhang2021gan, liu2022target, zhou2021semantic}. It is worth noting that the efficient integration of information extracted from different sources to enhance the representation capability of the fusion results is equally important.

To address these shortcomings, this paper proposes GAN-HA. Considering the weakness of the structurally identical dual-discriminator GAN methods, we purposefully design two distinct discriminators tailored to the specific features of infrared and visible images. The salient discriminator is designed for intensity information in infrared images and consists of a multi-scale channel attention (MS-CA) module and an infrared global discriminative (\(D_{ir}\)) module, the detailed discriminator is designed for the gradient information of the visible image and consists of a multi-scale spatial attention (MS-SA) module and a visible Markovian discriminative (\(D_{vi}\)) module. MS-CA and MS-SA help the discriminators focus more effectively on the important information of their respective source images, while \(D_{ir}\) and \(D_{vi}\) enable better learning of thermal radiation information and texture details, respectively. Besides, we design a new attention-based fusion strategy, which assigns corresponding weights to different features by calculating the differences of the corresponding pixels, thereby achieving effective fusion and appropriate information representation. As a result, our fused image obtains richer thermal radiation information and texture details and has better information representation capabilities. This paper demonstrates the model's superior performance through qualitative and quantitative experiments; its robustness through resistance to noise degradation tests; its high application potential through object detection and visual tracking experiments; and its generalization ability through extended biomedical fusion experiments. 

The contributions of this work are summarized below:
\vspace{-0.2cm}
\begin{itemize}
\item [(1)]
A novel heterogeneous dual-discriminator network is proposed for the first time. The salient discriminator, comprising MS-CA and \(D_{ir}\), facilitates the comprehensive learning of global thermal region, while the detail discriminator, consisting of MS-SA and \(D_{vi}\), excels in capturing local details. This innovative design adapts to the unique characteristics of infrared and visible images, enabling targeted and concurrent learning of both thermal radiation information and texture details.
\vspace{-0.3cm}
\item [(2)]
A new attention-based fusion strategy is proposed in the generator. After utilizing the disparities between source images to integrate the acquired image information, this approach ensures the fusion results effectively highlight the features learned from different sources.
\vspace{-0.3cm}
\item [(3)]
Comprehensive experiments are conducted on various public datasets. Qualitative and quantitative comparisons demonstrate the superior performance of GAN-HA. Moreover, experiments of noise degradation resistance, downstream object detection and visual tracking, and extended biomedical image fusion, further validate the significant potential of the approach for practical applications.
\end{itemize}

The subsequent sections of this paper are organized as follows. Section \ref{section:B} provides an overview of traditional model-based IVIF technology and deep learning-based IVIF technology. In section \ref{section:C}, we delve into the structure of the proposed model and detail its advantages. Section \ref{section:D} presents extensive experimental validations, including qualitative and quantitative analyses, ablation experiments, noise degradation resistance comparisons, object detection and visual tracking application evaluations, and biomedical image fusion experiments. Finally, conclusions are presented in section \ref{section:E}.

\section{Related works}
\label{section:B}
\vspace{-0.2cm}
\subsection{Model-Based IVIF}
\vspace{-0.1cm}
In model-based IVIF methods, feature extraction operators and fusion rules are designed to attain desired outcomes. As previously delineated, these methods can be classified into four primary categories. The first category is MST-based methods, which are grounded in the theory that images can be decomposed into different components based on the varying scales of different features. A satisfactory fusion can be achieved by integrating these components through specific design rules. Representative studies include pyramid transform \cite{du2016union}, wavelet transform \cite{lewis2007pixel}, and curvelet transform \cite{nencini2007remote}. The second category is SR methods. It has been observed that visual recognition systems process images in a sparse manner. Therefore, it is feasible to enhance the representation of valid information by learning an over-complete dictionary and characterizing image features with sparse coefficients \cite{zhang2018robust, wang2014fusion, liu2016image}. The third category is subspace-based methods. Recognizing the redundancy present in most images, some scholars propose to capture internal structural information by mapping images to a low-dimensional subspace to effectively eliminate interference from this redundant information \cite{mishra2014mri, cvejic2007region}. The fourth category is hybrid methods, which aim to improve fusion performance by combining the advantages of the above methods \cite{liu2015general, he2023strategic}. Additionally, novel perspectives have emerged in IVIF, such as GTF \cite{ma2016infrared} proposed by Ma et al. in 2016, based on the theory of total variation (TV). Unfortunately, traditional methods usually have difficulty in fully considering all aspects and details of the integration process, which may limit their application in some tasks.

\vspace{-0.2cm}
\subsection{Deep Learning-Based IVIF}
\vspace{-0.1cm}
Compared to model-based methods that rely on manually designed rules, deep learning-based methods leverage the powerful fitting capability of neural networks, which makes them superior and therefore has attracted great attention in the field of image fusion.

\setlength{\parindent}{0.8em}
\textit{1) CNN for IVIF:} 
The CNN methods utilize convolutional neural networks to extract features from images, enhancing the fusion process and improving image quality. Within the CNN-based framework, Liu et al. introduced a multi-focus fusion strategy in 2017 \cite{liu2017multi}, which facilitates fusion tasks by learning the mapping between source images and focus maps. However, this method adopts a simple architecture, leading to underutilization of features. Li et al. utilized a pre-trained ResNet-50 model to extract deep features from images in 2019 \cite{li2019infrared}. Although this approach uses appropriate post-processing to obtain fused images, the design of the post-processing is complex. Zhang et al. employed sequential convolutional layers to extract features and enrich fusion results with information in 2020 \cite{zhang2020ifcnn}. Nevertheless, since this method is designed as a general image fusion model, it does not achieve superior performance in IVIF.

\textit{2) AE for IVIF:} 
The AE methods usually pre-train an autoencoder network, and then apply it to learn the representation of images and reconstruct the original data. In 2018, Li et al. proposed a novel codec model \cite{li2018densefuse}, leveraging dense connections to enhance information utilization. However, this codec structure discards the down-sampling operator, which is unable to comprehensively learn the information of images. In 2020, Li et al. achieved feature reuse through a multi-scale approach by focusing on features at varying scales \cite{li2020nestfuse}. Nevertheless, the design of an appropriate fusion strategy has always been a challenge. In 2021, Wang et al. innovatively combined the strengths of these two AE models to simultaneously aggregate useful features from both horizontal and vertical directions \cite{wang2021unfusion}. However, the lack of suitable guide images for training limits its fusion performance.

\setlength{\parindent}{0.8em}
\textit{3) Dual-Discriminator GAN for IVIF:} 
The GAN methods learn to generate realistic image samples by utilizing source images to guide the model during training. Ma et al. proposed DDcGAN \cite{ma2020ddcgan} in 2020, which contains two discriminators with identical structures to force the model to learn detailed features and thermal radiation information separately. Unfortunately, due to the limitations of the structure, some details are lost during information extraction. In 2021, Li et al. introduced AttentionFGAN \cite{li2020attentionfgan}. They incorporated a multi-scale attention structure in the generator and discriminators to better capture salient information. However, this attention structure often leads to the loss of some important texture features and boundary information. In the same year, Zhang et al. proposed GAN-FM \cite{zhang2021gan} to learn richer information through improved discriminator structure. Unlike the traditional global discriminators, it integrates a Markovian discriminative structure to better focus on local block structures. However, this discriminator structure lacks attention to intensity information. In 2022, Liu et al. presented TarDAL \cite{liu2022target}, which employs a joint training strategy for downstream detection tasks to achieve a bilevel optimization between fusion and detection. In 2023, Zhou et al. developed SDDGAN \cite{zhou2021semantic}, a semantic-supervised fusion network that can preserves the features of crucial semantic objects in images. Unfortunately, both TarDAL and SDDGAN have a similar structure to DDcGAN, and this simple structure is insufficient for adequately learning complex features. Therefore, existing dual-discriminator GAN methods overlook the limitations imposed by the structure, especially the discriminator structure. By employing a unified structure for both discriminators, these models struggle to simultaneously target the learning of intensity information in infrared images and gradient information in visible images.

\begin{figure*}[!t]
\setlength{\abovecaptionskip}{0.1cm}
\setlength{\belowcaptionskip}{-0.3cm}
  \centering
  \includegraphics[width=\linewidth]{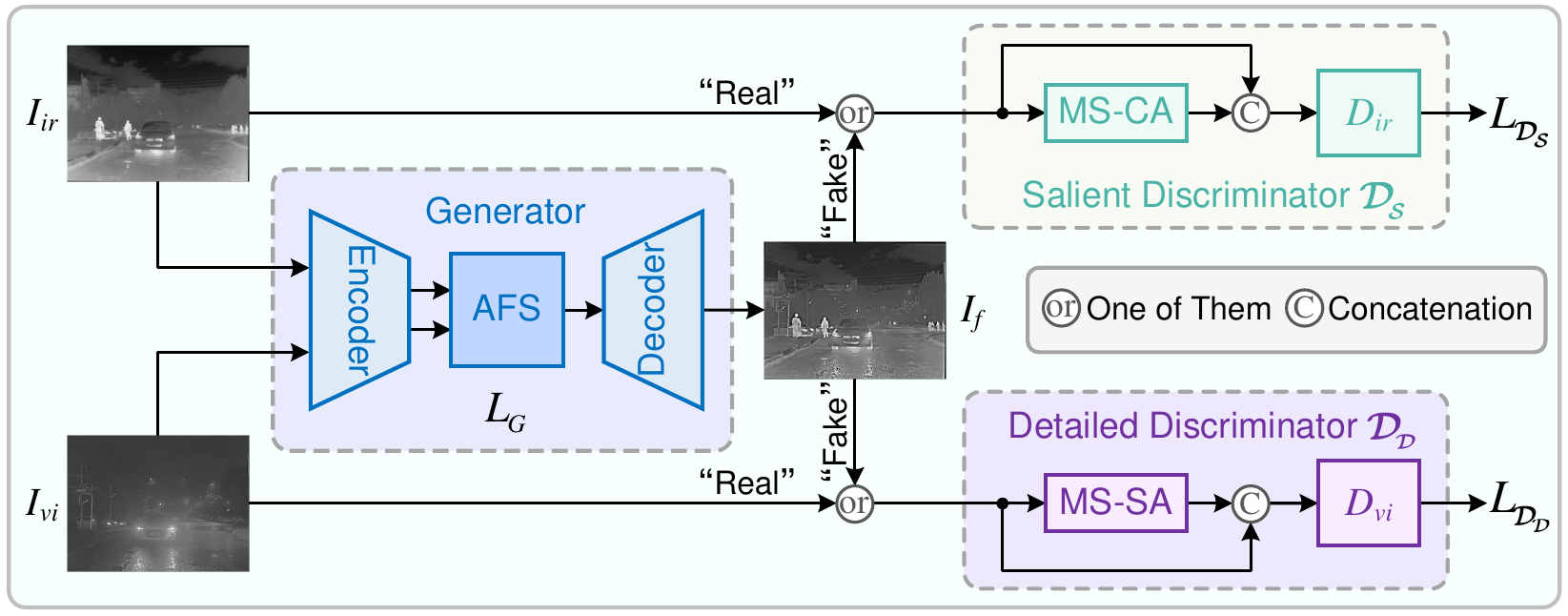}
  \caption{The framework of the proposed GAN-HA.}
  \label{Fig.2}
\end{figure*}

\section{Method}
\label{section:C}
\vspace{-0.2cm}
\subsection{Framework Overview}
\vspace{-0.1cm}
In this paper, a novel GAN-HA is proposed, and the overall framework is shown in Fig. \ref{Fig.2}. This model comprises two networks: the generator network and the dual-discriminator network. The generator is designed as a multi-scale skip-connected codec structure to realize information extraction, fusion and image reconstruction. Two discriminators are defined as different structures to learn thermal region and texture detail information, respectively, which are the salient discriminator targeting infrared images and the detailed discriminator focusing on visible images.

\begin{figure}[t]
\setlength{\abovecaptionskip}{0.1cm}
\setlength{\belowcaptionskip}{-0.5cm}
  \centering
  \includegraphics[width=\linewidth]{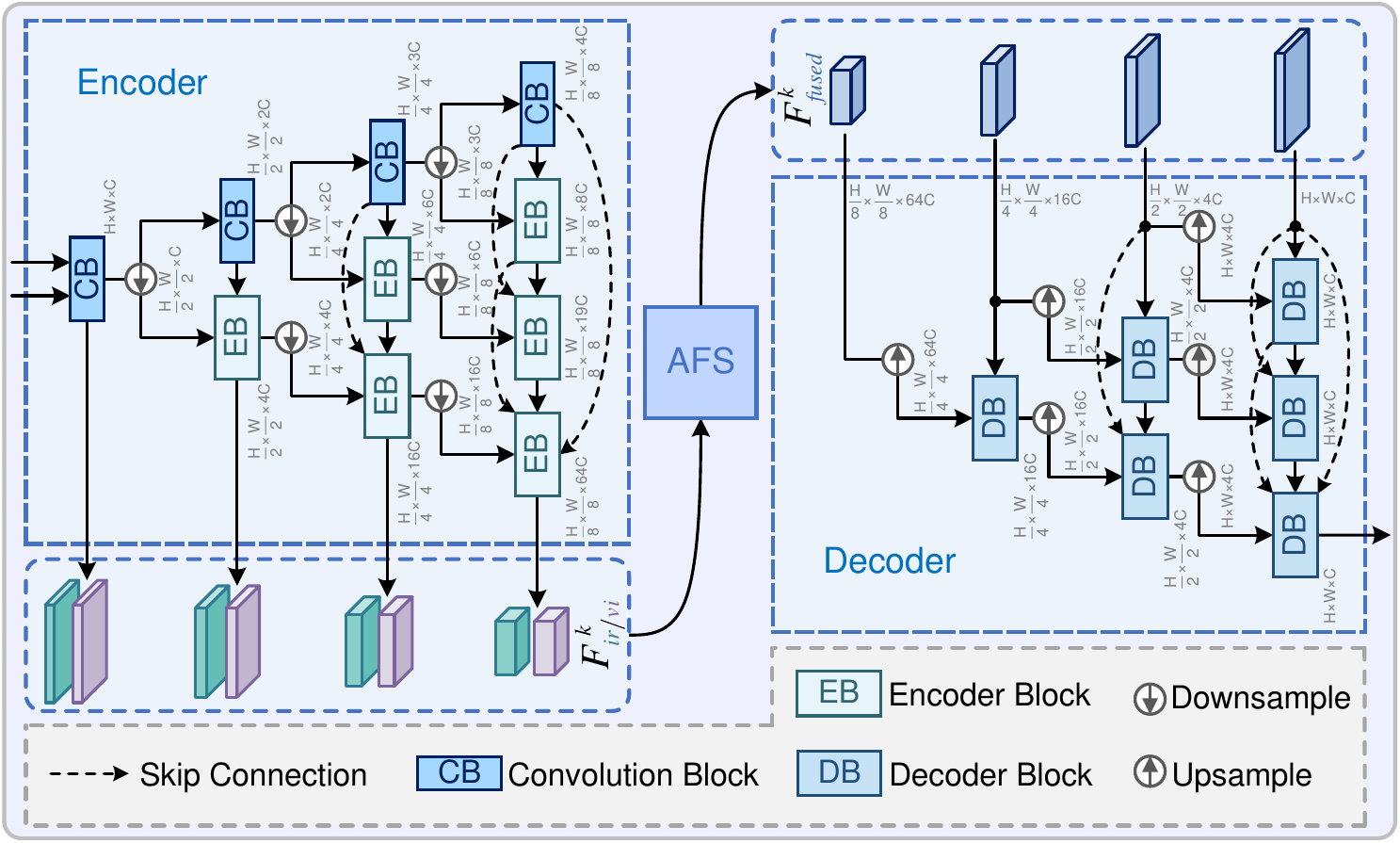}
  \caption{Detailed structure of the generator.}
  \label{Fig.3}
\end{figure}

\vspace{-0.2cm}
\subsection{Architecture of Generator}
\vspace{-0.1cm}
Inspired by Nestfuse \cite{li2020nestfuse}, we design the generator as a multi-scale skip-connected structure to effectively utilize image information. As shown in Fig. \ref{Fig.3}, the generator consists of three sub-networks: encoder, attention-based fusion strategy (AFS), and decoder. 

\setlength{\parindent}{0.8em}
\textit{1) Encoder and Decoder in Generator:} 
The encoder and decoder sub-networks mainly consist of the convolutional block (CB), encoder block (EB), and decoder block (DB). CB contains one convolutional layer, EB denotes two convolutional operations with different kernel sizes, and DB consists of two identical convolutional layers. Specifically, after importing the source images \(I_{ir}\) and \(I_{vi}\) into the encoder sub-network, deep feature maps \(F_{ir}^k\) and \(F_{vi}^k\) with different sizes are obtained respectively through the CB, EB, and down-sampling operations. Here \textit{k} represents the scale level of the feature maps. Then, after traversing AFS, the fused deep feature \(F_{fused}^k\) is produced. Ultimately, the final fused image \(I_{f}\) is generated after a series of up-sampling and DB operations in the decoder sub-network. The sampling operations within the network are used to exploit multi-scale features, while the skip connections are used to enhance the extraction and preservation of intermediate features.

\begin{figure}[t]
\setlength{\abovecaptionskip}{0.1cm}
\setlength{\belowcaptionskip}{-0.4cm}
  \centering
  \includegraphics[width=\linewidth]{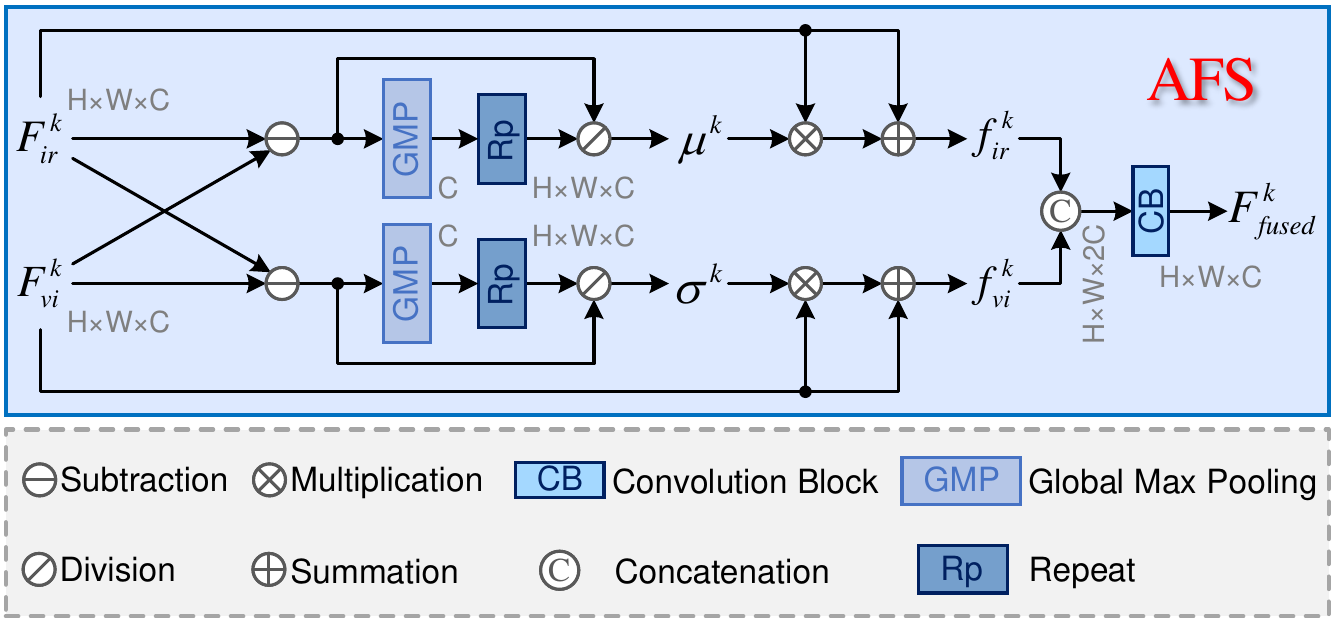}
  \caption{Construction of AFS.}
  \label{Fig.4}
\end{figure}

\setlength{\parindent}{0.8em}
\textit{2) AFS in Generator:} 
\label{section:C3}
As previously mentioned, effectively fusing information obtained from different source images is crucial. In particular, dual-discriminator GANs may face the challenge related to learning imbalance. Therefore, this paper designs AFS as the information fusion layer. AFS integrates the information reasonably and effectively to enhance the information representation ability of the fusion result. The detailed structure is shown in Fig. \ref{Fig.4}.

From the regions framed in Fig. \ref{Fig.1}, it can be observed that the greater the differences between corresponding regions of the source images, the more important the specific information of the corresponding source images becomes. Therefore, it is logical to represent the different importance of these information based on the distance between the source images, and then assign corresponding weights accordingly. The formula for calculating the weighting coefficients is as follows:

\begin{equation}
{\mu ^k}  = \frac{{F_{ir}^k - F_{vi}^k}}{{Rp\left( {GMP\left( {F_{ir}^k - F_{vi}^k} \right)} \right)}}
\end{equation}
\begin{equation}
{\sigma ^k} = \frac{{F_{vi}^k - F_{ir}^k}}{{Rp\left( {GMP\left( {F_{vi}^k - F_{ir}^k} \right)} \right)}}
\end{equation}
where \(Rp\left ( \cdot  \right ) \) and \(GMP\left ( \cdot  \right ) \) donate repeat and global max pooling, respectively. Both \({\mu ^k}\) and \({\sigma ^k}\) reflect the importance of different information in feature maps.

Then, the feature maps that can highlight the respective important information are calculated through the following formulas:
\begin{equation}
f_{ir}^k = {\mu ^k} * F_{ir}^k + F_{ir}^k
\end{equation}
\begin{equation}
f_{vi}^k = {\sigma ^k} * F_{vi}^k + F_{vi}^k
\end{equation}
where \(\ast \) is the multiplication operation. 

Finally, the fused features \(F_{fused}^k\) are obtained after passing through the concatenation operation (\(concat\left ( \cdot  \right ) \)) and the convolution layer (\(conv\left ( \cdot  \right ) \)):
\begin{equation}
F_{fused}^k = conv\left( {concat\left( {f_{ir}^k,f_{vi}^k} \right)} \right)
\end{equation}

The final deep feature maps \(F_{fused}^k\) integrate the information from different source images reasonably, effectively alleviating the problem of unbalanced information representation in the fusion results.

\begin{figure*}[!t]
\setlength{\abovecaptionskip}{0.1cm}
\setlength{\belowcaptionskip}{-0.1cm}
  \centering
  \includegraphics[width=\textwidth]{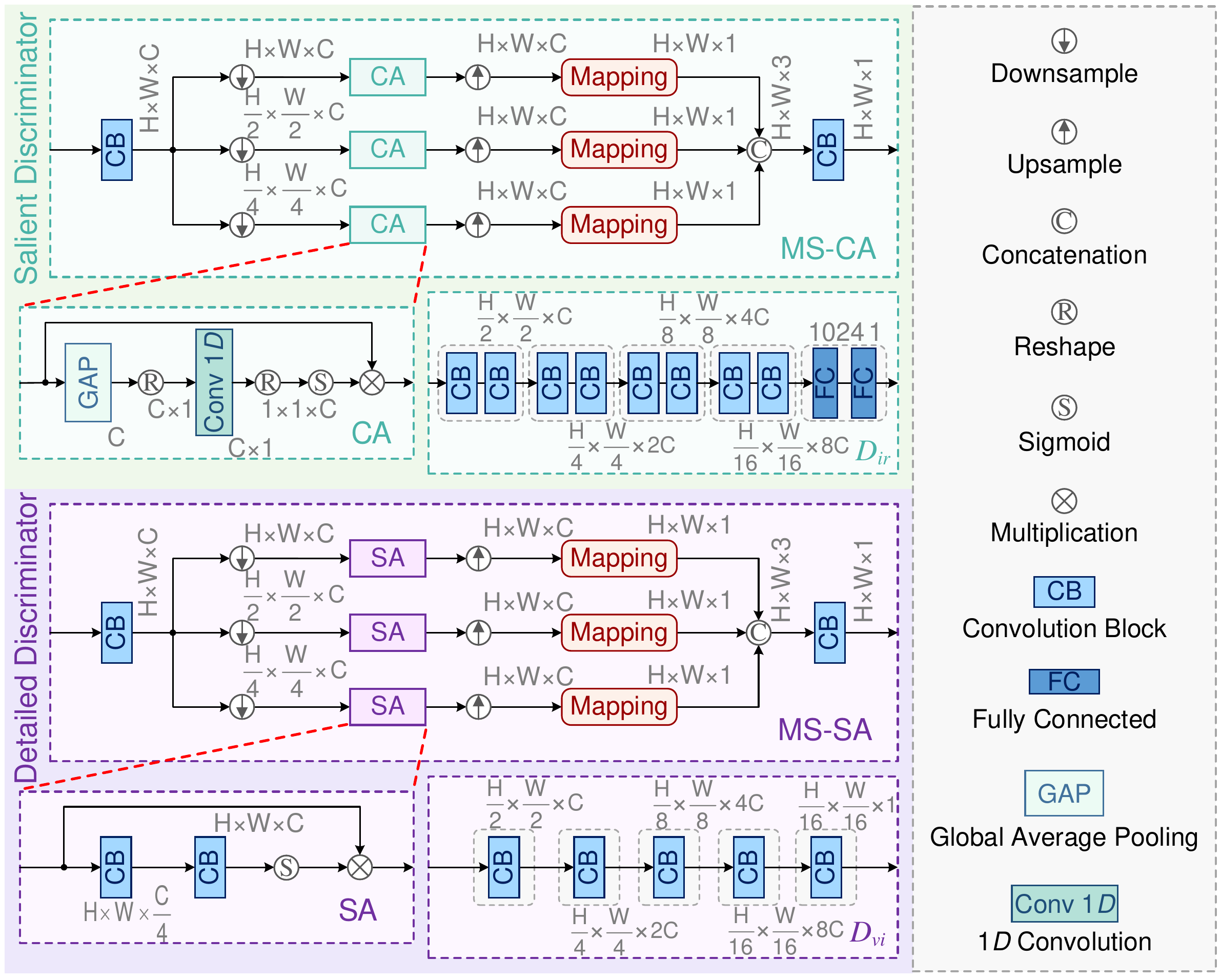}
  \caption{Details of discriminator networks.}
  \label{Fig.5}
\end{figure*}

\vspace{-0.2cm}
\subsection{Architecture of Discriminators}
\vspace{-0.1cm}
In GAN-HA, the dual-discriminator network consists of two parts: the salient discriminator (\({\mathcal{D}_\mathcal{S}}\)) and the detailed discriminator (\({\mathcal{D}_\mathcal{D}}\)). The specific architecture is displayed in Fig. \ref{Fig.5}. The input images are first fed into the multi-scale channel attention (MS-CA) module and multi-scale spatial attention (MS-SA) module respectively to calculate the corresponding attention maps. After that, these maps are concatenated with input images accordingly and fed into the corresponding infrared global discriminative (\(D_{ir}\)) module and visible Markovian discriminative (\(D_{vi}\)) module for discrimination.

\setlength{\parindent}{0.8em}
\textit{1) Salient Discriminator:} 
To facilitate the model to focus on learning thermal region information from infrared images during adversarial training, \({\mathcal{D}_\mathcal{S}}\) is split into two components: the MS-CA module and the \(D_{ir}\) module. 

The MS-CA structure is introduced to constrain the discriminator to concentrate on the most critical thermal regions in infrared images. First, since single-scale features may not capture all essential thermal radiation information, the multi-scale mechanism is adopted to utilize more deep features via down-sampling. Additionally, to focus on the distribution of the overall pixels in images, i.e., the intensity information of the thermal regions, the channel attention (CA) module \cite{wang2020eca} is designed to reweight the global information of the infrared features through global average pooling (GAP) and one-dimensional convolution. Afterward, the up-sampling operation is used to restore images to their original size, and the mapping operation (selecting the maximum value of all features in channel dimensions \cite{zhou2016learning}) is used to calculate weighted maps. Finally, these images are concatenated to form the ultimate attention maps. 

The role of \(D_{ir}\) is to distinguish between the fused output and the original infrared image, and it consists of five layers. The first four layers are all two identical convolution operations and the last layer includes two fully connected operations. After passing through this module, the result of the discrimination (the final determine probability of the input image) is generated. 

\setlength{\parindent}{0.8em}
\textit{2) Detailed Discriminator:}
\({\mathcal{D}_\mathcal{D}}\) can help preserve texture details more effectively during adversarial learning. 

The aim of MS-SA is to better highlight the information of important local details. The included spatial attention (SA) module \cite{liu2021global} can capture more relationships between pixel spaces by decreasing and increasing the number of channels during two convolutions, thereby giving greater weight to critical spatial structure information. 

\(D_{ir}\) is adopted to encourage the model to learn the detailed information of visible images fully. It can focus on local details by classifying each block of the input image \cite{isola2017image}. As shown in Fig. \ref{Fig.5}. \(D_{ir}\) consists of five convolution layers, and the output is a matrix that can be averaged to obtain the final determine probability.

\vspace{-0.2cm}
\subsection{Loss Function}
\label{section:C5}
\vspace{-0.1cm}
As depicted in Fig. \ref{Fig.2}, the loss function is comprised of two parts: the loss function of the generator (\(L_{G}\)), and the loss function of discriminators (\(L_{\mathcal{D}_\mathcal{S}}\) and \(L_{\mathcal{D}_\mathcal{D}}\)).

\setlength{\parindent}{0.8em}
\textit{1) Loss Function of Generator:}
In order to force the generated image to contain sufficient information from both the infrared and visible images, the loss function used to guide the training of the generator consists of the adversarial loss (\(L_{adver}\)) and the basic loss (\(L_{basic}\)):
\begin{equation}
\label{Eq.6}
L_{G} = L_{adver} + \alpha L_{basic}
\end{equation}
where \(\alpha\) is the parameter to control the two terms.

\noindent\textbf{Adversarial Loss \(\bm{L_{adver}}\).}
Considering that the generator needs to deceive both discriminators during training, the adversarial loss can be defined as follows:
\begin{equation}
\begin{split} 
{L_{{adver}}} = \mathbb{E} {_{{I_f} \sim {d_{{I_f}}}}} & \left [ \left( {\mathcal{D}_\mathcal{S}\left( {{I_f}} \right) - \phi } \right)^{2}  \right ] 
\\ & +  \mathbb{E}  {_{{I_f} \sim {d_{{I_f}}}}} \left  [ \left( {\mathcal{D}_\mathcal{D}\left( {{I_f}} \right) - \phi } \right)^{2}  \right ]
\end{split}
\end{equation}
where \(\mathbb{E}\) represents the mathematical expectation, \({I_f}\) denotes the fused image and \({d_{{I_f}}}\) is its distribution, \(\phi\) refers to a probability label (the discriminative probability of discriminators for the generated image) and is set to 1 because the generator expects the discriminators to be unable to distinguish between images.

\noindent\textbf{Basic Loss \(\bm{L_{basic}}\).}
The purpose of \(L_{basic}\) is to constrain the learning of valid information in source images, and is divided into two items: the learning of thermal radiation information and the learning of texture details:
\begin{equation}
\label{Eq.8}
L_{basic} = L_{visible} + \beta L_{infrared}
\end{equation}
The equations for \({L_{infrared}}\) and \({L_{visible}}\) are as follows:
\begin{equation}
{L_{infrared}} = \left\| {{I_f} - {I_{ir}}} \right\|_F^2
\end{equation}
\begin{equation}
{L_{visible}} = {\left\| {\nabla \left( {{I_f}} \right) - \nabla \left( {{I_{vi}}} \right)} \right\|_1}
\end{equation}
where \(\beta\) is the trade-off parameter,  \(\nabla \left(\cdot  \right)\) is the Sobel gradient operator known for its better ability to resist interference, \(\left\| {\cdot } \right\|_F^2\) and \({\left\| {\cdot } \right\|_1}\) denote the Frobenius norm and \({\ell_1}\) norm.

\setlength{\parindent}{0.8em}
\textit{2) Loss Function of Discriminators:}
The loss function of discriminators is composed of the loss of the salient discriminator \(L_{\mathcal{D}_\mathcal{S}}\) and the loss of the detailed discriminator \(L_{\mathcal{D}_\mathcal{D}}\). Since GAN-HA has two heterogeneous discriminators that may aggravate the imbalance conflict when the generator learns different features, \(\gamma\) is introduced to allow the model to adjust the optimization \cite{li2020attentionfgan}:
\begin{equation}
\label{Eq.11}
L_{D} = L_{\mathcal{D}_\mathcal{D}} + \gamma L_{\mathcal{D}_\mathcal{S}}
\end{equation}
The equations for \({L_{\mathcal{D}_\mathcal{S}}}\) and \({L_{\mathcal{D}_\mathcal{D}}}\) are as follows:
\begin{equation}
\begin{split} 
{L_{\mathcal{D}_\mathcal{S}}}  = \frac{1}{2} \times \mathbb{E} {_{{I_{ir}} \sim {d_{I_{ir}}}}} & \left [\left({\mathcal{D}_\mathcal{S}\left( {{I_{ir}}}\right)-\xi_{1} } \right)^{2}\right] 
\\  + \frac{1}{2} \times & \mathbb{E}  {_{{I_f} \sim {d_{{I_f}}}}} \left  [ \left( {\mathcal{D}_\mathcal{S}\left( {{I_f}} \right) - \xi_{2} } \right)^{2}  \right ]
\end{split}
\end{equation}
\begin{equation}
\begin{split} 
{L_{\mathcal{D}_\mathcal{D}}} = \frac{1}{2} \times \mathbb{E} {_{{I_{vi}} \sim{d_{I_{vi}}}}} & \left [\left({\mathcal{D}_\mathcal{D}\left({{I_{vi}}}\right)-\zeta_{1}}\right)^{2}\right] 
\\ + \frac{1}{2} \times & \mathbb{E}  {_{{I_f} \sim {d_{{I_f}}}}} \left  [ \left( {\mathcal{D}_\mathcal{D}\left( {{I_f}} \right) - \zeta_{2} } \right)^{2}  \right ]
\end{split}
\end{equation}
where \({I_{ir}}\) denotes the infrared image and \({d_{{I_{ir}}}}\) is its distribution, while \({I_{vi}}\) denotes the visible image and \({d_{{I_{vi}}}}\) is its distribution. The probability labels \(\xi_{1}\) and \(\xi_{2}\) are set to 1 and 0, respectively, and \(\zeta_{1}\) and \(\zeta_{2}\) are similarly set to 1 and 0, respectively. This is because the discriminators expect to distinguish the input image as accurately as possible (giving the probability value of the real image as large as possible as well as the probability value of the generated image as small as possible).

\vspace{-0.2cm}
\subsection{Discussion}
\vspace{-0.1cm}
Since the utilization of heterogeneous discriminant structures has not been considered in previous dual-discriminator models, we justify the proposed GAN-HA in the following three aspects:

\setlength{\parindent}{0.8em}
\textit{\textbf{1.}}
Considering that what needs to be learned from infrared images is intensity information, while what needs to be learned from visible images is gradient information. It is logical to design two heterogeneous discriminators, respectively, to learn the features of different source images. This study is intended to bring some inspiration.

\setlength{\parindent}{0.8em}
\textit{\textbf{2.}}
The rationality of GAN-HA is evident in the overall performance of the model and the effectiveness of the key heterogeneous discriminator structures. Following the experimental content referenced in \cite{zhang2021image,he2023reti}, the qualitative and quantitative experiments are conducted in subsection \ref{section:D1} to evaluate the overall impact of the model. The ablation experiments on heterogeneous discriminator structures are carried out in subsection \ref{section:D2} for evaluation.

\setlength{\parindent}{0.8em}
\textit{\textbf{3.}}
The rationality of GAN-HA is also reflected in its application potential and generalization capability. Therefore, resistance to noise degradation is compared in section \ref{section:D3}. Downstream object detection and tracking experiments are carried out in section \ref{section:D4}. Extended fusion experiments in the biomedical field are performed in section \ref{section:D5}.

\section{Experiment}
\label{section:D}
\vspace{-0.2cm}
\subsection{Implementation Details}

\setlength{\parindent}{0.8em}
\textit{1) Datasets:}
The experiments are conducted on three public datasets: RoadScene \cite{xu2020u2fusion}, LLVIP \cite{jia2021llvip}, and \(M^3FD\) \cite{liu2022target}. The RoadScene dataset includes 221 pairs of infrared-visible images depicting various traffic scenes. LLVIP contains traffic images captured in low-light conditions and we select 300 pairs for our experiments. \(M^3FD\) is a newly available multi-scene multimodal dataset consisting of 300 image pairs. We randomly divide the datasets into training and testing sets. The details are listed in Table \ref{Tab.1}. Subsequently, we uniformly crop the training set into image blocks of size 
 128 \(\times\) 128 to augment the training images, resulting in a total of 13,903 images.

\begin{table}[t]
\setlength{\abovecaptionskip}{0.1cm}
    \caption{Details of datasets division.}
    \centering
    \tabcolsep=0.15cm
    \renewcommand\arraystretch{1}
    \label{Tab.1}
    \begin{tabular}{c|cc|c}
        \toprule[0.5mm]
        Datasets &Training Set&Testing Set&Total\\ \hline\hline
        RoadScene&181         &40        &221   \\
        LLVIP    &100         &200        &300  \\
        \(M^3FD\)&100         &200        &300  \\ 
        \toprule[0.5mm]
    \end{tabular}
\vspace{-0.5cm}
\end{table}

\setlength{\parindent}{0.8em}
\textit{2) Comparative Methods:}
Our method, GAN-HA, is compared with nine SOTA methods, including a typical traditional method, GTF \cite{ma2016infrared}, and eight deep learning-based methods. U2Fusion \cite{xu2020u2fusion} and STDFusion \cite{ma2021stdfusionnet} are CNN-based fusion models, while DeepFuse \cite{ram2017deepfuse}, AUIF \cite{zhao2021efficient}, and DRF \cite{xu2021drf} utilize AE-based networks. In addition, FusionGAN \cite{ma2019fusiongan}, GANMcC \cite{ma2020ganmcc}, and GAN-FM \cite{zhang2021gan} are GAN-based methods. All comparison methods have publicly available code, and we refer to the corresponding original papers for parameter setting, retraining and testing.

\setlength{\parindent}{0.8em}
\textit{3) Evaluation Metric:}
Following \cite{xu2023dm, he2023degradation}, we use six commonly used metrics to evaluate the performance of IVIF methods from multiple perspectives, including entropy (EN) \cite{roberts2008assessment}, average gradient (AG) \cite{li2021different}, spatial frequency (SF) \cite{gioux2019spatial}, feature mutual information (FMI) \cite{haghighat2011non}, visual information fidelity (VIF) \cite{han2013new}, and universal image quality index (UIQI) \cite{wang2002universal}. EN measures the information richness of the fused image; AG evaluates the quality based on gradients; SF reflects the rate of grayscale change in the image; FMI calculates the mutual information of image features; VIF assesses the visual quality of the fused image; And UIQI compares the image brightness. In summary, larger values for these metrics indicate better fusion effects.

\setlength{\parindent}{0.8em}
\textit{4) Training Details:}
The TensorFlow and Adam optimizer are used in our method. During the training process, the generator and discriminators are updated alternately and iteratively. Specifically, after training \({\mathcal{D}_\mathcal{D}}\) four times and training \({\mathcal{D}_\mathcal{S}}\) twice, the generator is trained twice. The epochs and batch sizes are set to 20 and 16, respectively, and the initial learning rate is \(2\times 10^{-4}\). Except for GTF, which is executed on MATLAB R2023a, all other methods run on an NVIDIA RTX 2080 Ti GPU. Additionally, the hyperparameters \(\alpha\), \(\beta\), and \(\gamma\) in Equations \ref{Eq.6}, \ref{Eq.8}, and \ref{Eq.11} are set to 100, 5, and 5, respectively.

\vspace{-0.2cm}
\subsection{Results and Analysis on Public Datasets}
\label{section:D1}
\vspace{-0.1cm}

\begin{figure*}[t]
\setlength{\abovecaptionskip}{0cm}
\setlength{\belowcaptionskip}{-0.4cm}
  \centering
  \includegraphics[width=\textwidth]{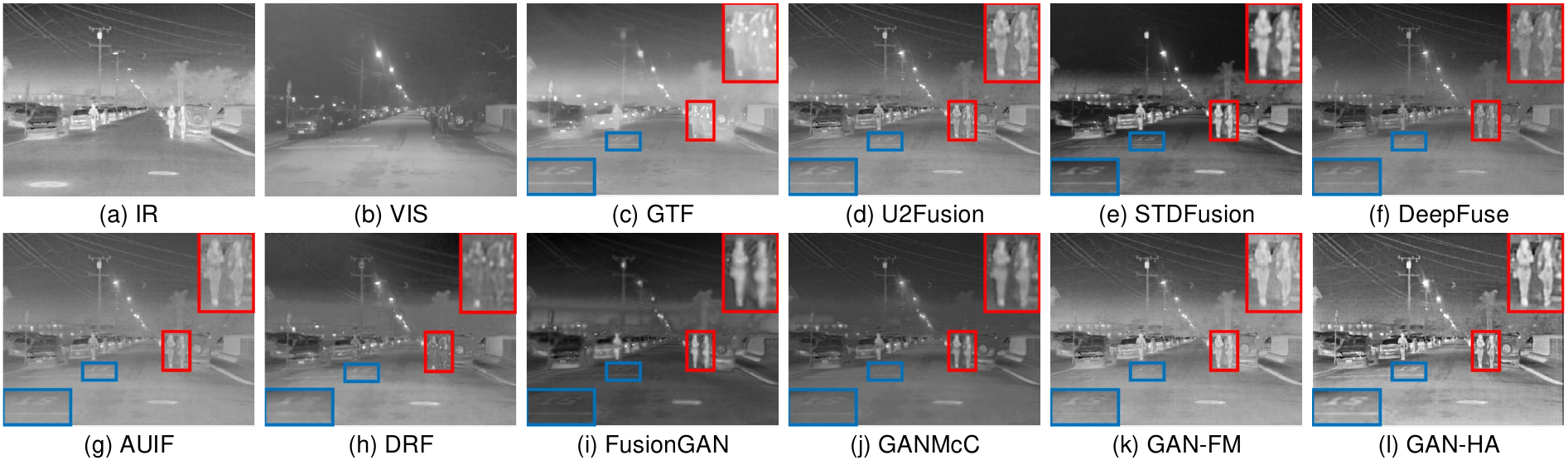}
  \caption{Qualitative results for RoadScene. (a) and (b) are the infrared image and visible image. (c)-(l) are the fused images processed by the ten comparison methods.}
  \label{Fig.6}
\end{figure*}

\begin{figure*}[!t]
\setlength{\abovecaptionskip}{0cm}
\setlength{\belowcaptionskip}{-0.4cm}
  \centering
  \includegraphics[width=\textwidth]{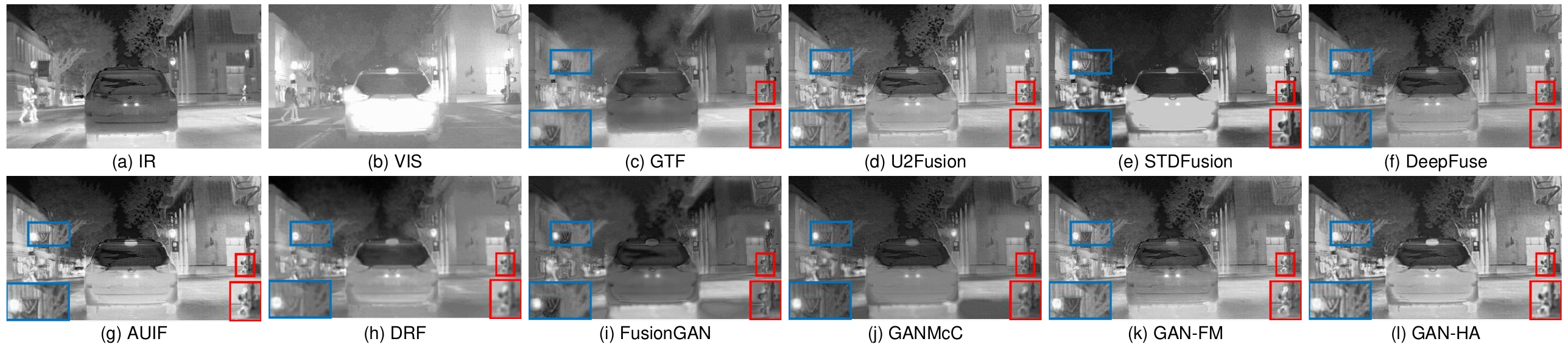}
  \caption{Qualitative results for RoadScene. (a) and (b) are the infrared image and visible image. (c)-(l) are the fused images processed by the ten comparison methods.}
  \label{Fig.7}
\end{figure*}

\begin{figure*}[!t]
\setlength{\abovecaptionskip}{0cm}
\setlength{\belowcaptionskip}{-0.4cm}
  \centering
  \includegraphics[width=\textwidth]{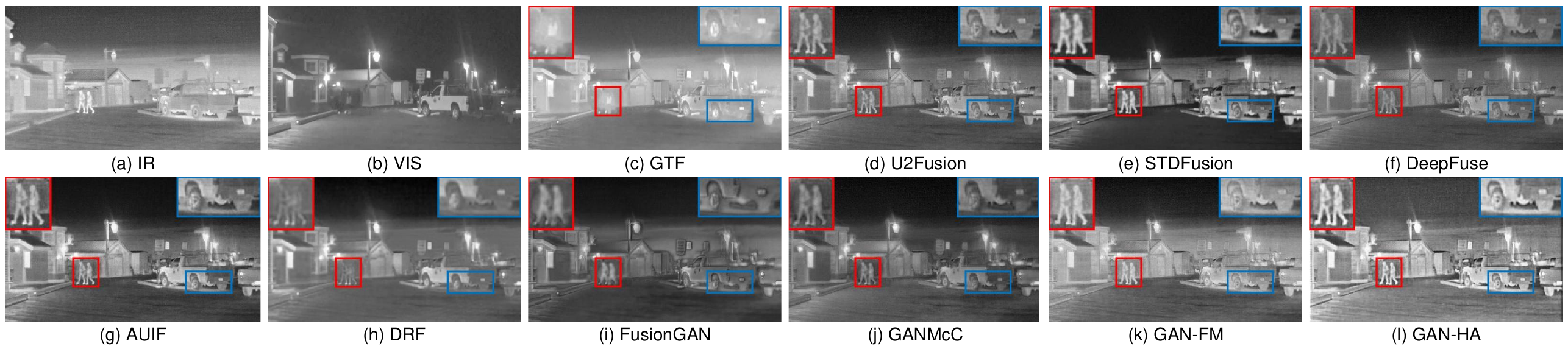}
  \caption{Qualitative results for RoadScene. (a) and (b) are the infrared image and visible image. (c)-(l) are the fused images processed by the ten comparison methods.}
  \label{Fig.8}
\end{figure*}

\begin{figure*}[t]
\setlength{\abovecaptionskip}{0cm}
\setlength{\belowcaptionskip}{-0.4cm}
  \centering
  \includegraphics[width=\textwidth]{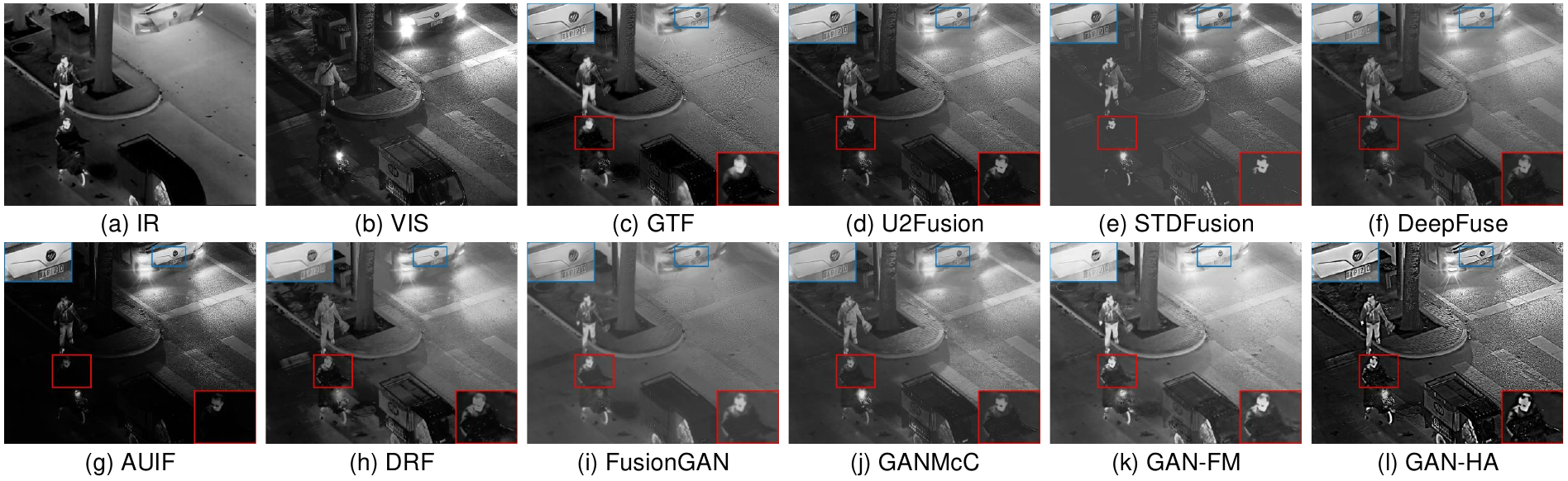}
  \caption{Qualitative results for LLVIP. (a) and (b) are the infrared image and visible image. (c)-(l) are the fused images processed by the ten comparison methods.}
  \label{Fig.9}
\end{figure*}

\begin{figure*}[t]
\setlength{\abovecaptionskip}{0cm}
\setlength{\belowcaptionskip}{-0.4cm}
  \centering
  \includegraphics[width=\textwidth]{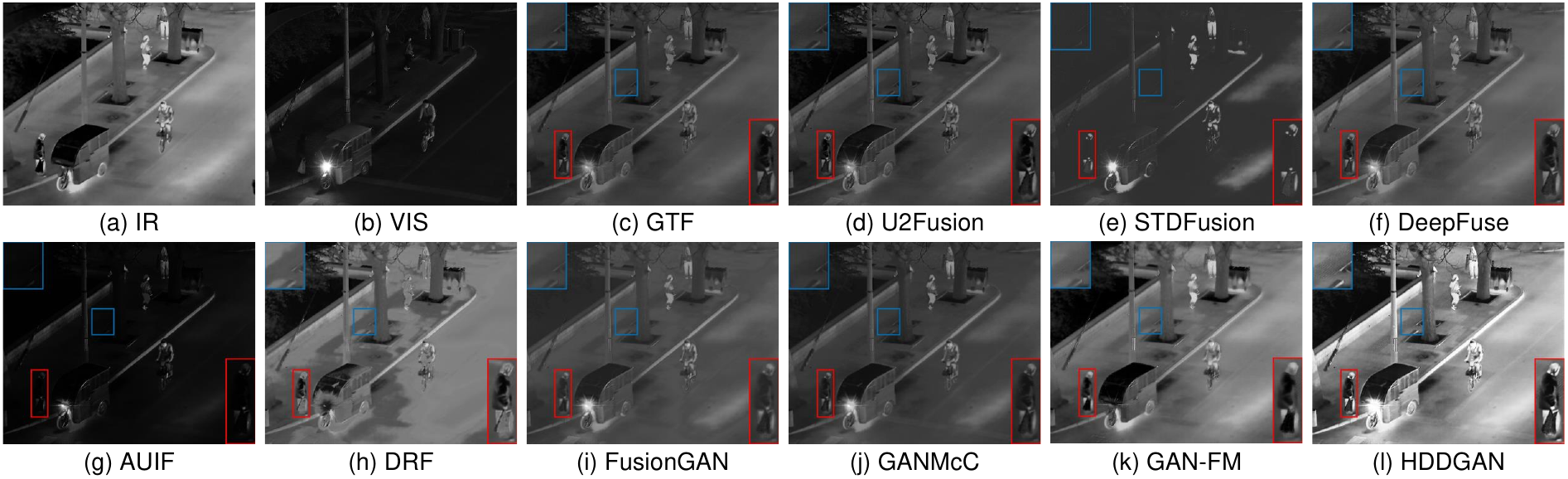}
  \caption{Qualitative results for LLVIP. (a) and (b) are the infrared image and visible image. (c)-(l) are the fused images processed by the ten comparison methods.}
  \label{Fig.10}
\end{figure*}

\begin{figure*}[t]
\setlength{\abovecaptionskip}{0cm}
\setlength{\belowcaptionskip}{-0.4cm}
  \centering
  \includegraphics[width=\textwidth]{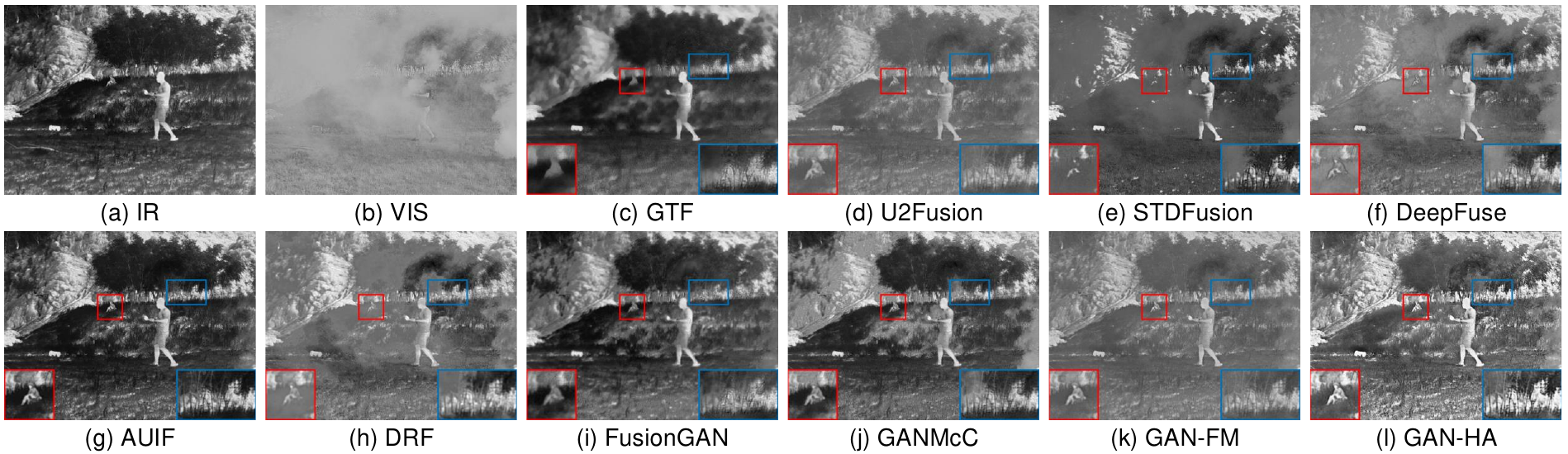}
  \caption{Qualitative results for $M^{3}FD$. (a) and (b) are the infrared image and visible image. (c)-(l) are the fused images processed by the ten comparison methods.}
  \label{Fig.11}
\end{figure*}

\begin{figure*}[!t]
\setlength{\abovecaptionskip}{0cm}
\setlength{\belowcaptionskip}{-0.4cm}
  \centering
  \includegraphics[width=\textwidth]{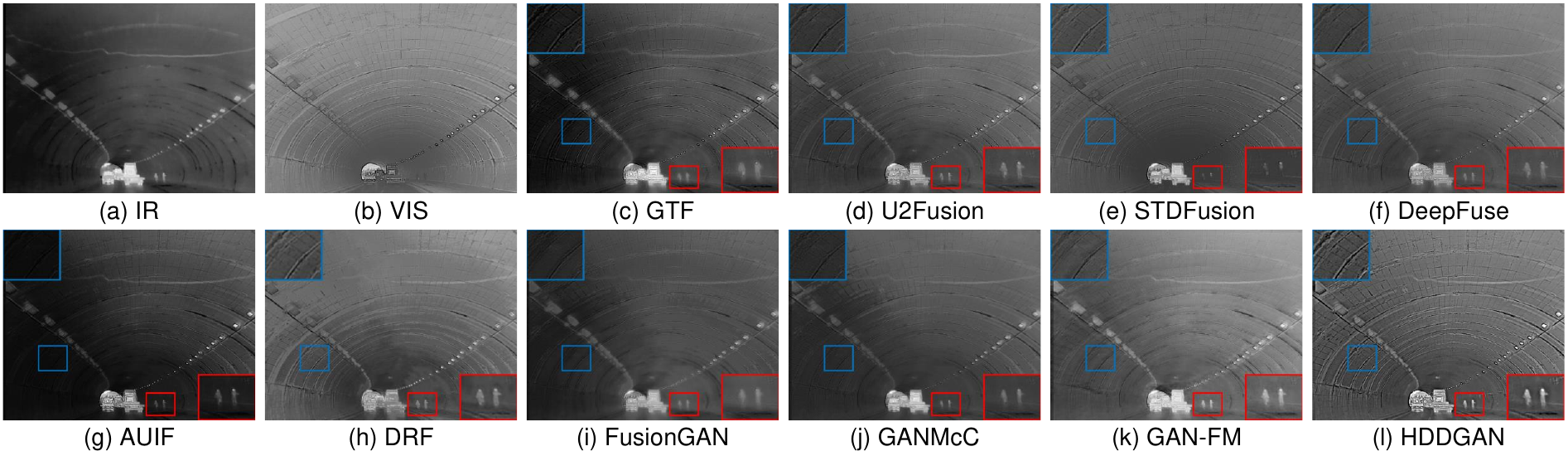}
  \caption{Qualitative results for $M^{3}FD$. (a) and (b) are the infrared image and visible image. (c)-(l) are the fused images processed by the ten comparison methods.}
  \label{Fig.12}
\end{figure*}

\setlength{\parindent}{0.8em}
\textit{1) Qualitative Comparisons:}
To demonstrate the superiority of GAN-HA, we qualitatively evaluate seven representative image pairs from the testing set and compare them with nine SOTA methods mentioned earlier. The fusion results are shown in Fig. \ref{Fig.6} - \ref{Fig.12}. To show the differences clearly, we zoom in on the thermal targets in the red-framed region and the texture details in the blue-framed region, and place them in one corner of the picture.

From a visual perspective, our method has several unique advantages. Firstly, the results of GAN-HA show similar pixel distributions to infrared images, while preserving texture information of visible images. This is closely aligned with the requirements of IVIF. Secondly, compared to other SOTA algorithms, GAN-HA excels in capturing more salient thermal region information and clearer detail information,  such as pedestrians in the red boxes in Fig. \ref{Fig.6}, \ref{Fig.8} - \ref{Fig.10}, and the details in the blue boxes in Fig. \ref{Fig.7} - \ref{Fig.12}. This advantage is attributed to the distinct structure of two discriminators, which allows the model to learn more from source images. Thirdly, GAN-HA could emphasize critical information beneficial for practical tasks. This is owed to the addition of MS-CA and MS-SA structures in discriminators and AFS in the generator. It can be seen that the visibility of hidden persons in Fig. \ref{Fig.7}, \ref{Fig.11}, \ref{Fig.12}, as well as the visibility of the lane lines and license plates in Fig. \ref{Fig.6} and Fig. \ref{Fig.9}, have been enhanced.

Furthermore, among the nine SOTA methods used for comparison, GTF tends to overlook boundary information (see Fig. \ref{Fig.6}). U2Fusion is too balanced to significantly enhance either thermal region or texture information (see Fig. \ref{Fig.11}). STDFusion and AUIF perform poorly in low-light conditions (see Fig. \ref{Fig.9}). DeepFuse, DRF, and  GANMcC typically lose some important texture features of visible images (see Fig. \ref{Fig.6}). Additionally, FusionGAN and GAN-FM consistently fail to retain high-contrast characteristics similar to infrared images (see Fig. \ref{Fig.12}).

\begin{table*}[t]
\setlength{\abovecaptionskip}{0.1cm}
    \centering
    \renewcommand\arraystretch{1.35}
    \caption{The average scores of the six metrics in RoadScene, LLVIP, and $M^{3}FD$ datasets, with the best two results are in \textcolor{red}{red} and \textcolor[rgb]{0,0.4470,0.7410}{blue}.}
    \resizebox{\linewidth}{!}{
    \begin{tabular}{p{1.8cm}<{\centering}|p{1.8cm}<{\centering}|p{1.8cm}<{\centering}|p{1.8cm}<{\centering}|p{1.9cm}<{\centering}|p{1.8cm}<{\centering}|p{1.8cm}<{\centering}|p{1.8cm}<{\centering}|p{2.0cm}<{\centering}|p{1.8cm}<{\centering}|p{1.8cm}<{\centering}|p{1.8cm}<{\centering}}
        \toprule[0.5mm]
        Datasets & Metrics & GTF & U2Fusion & STDFuison & DeepFuse & AUIF & DRF & FusionGAN & GANMcC & GAN-FM & GAN-HA \\ \hline\hline
        \multirow{6}{*}{RoadScene} & EN $\uparrow$ & \textcolor{red}{7.499} & 7.192  & 7.371  & 7.141  & \textcolor[rgb]{0,0.4470,0.7410}{7.472} & 7.111 & 7.048 & 7.209 & 7.431  & 7.293  \\
         ~ & AG $\uparrow$ & 3.529 & 5.084  & \textcolor[rgb]{0,0.4470,0.7410}{5.149} & 5.102  & 5.002  & 3.550 & 3.409 & 3.818 & 4.801 & \textcolor{red}{5.173} \\
         ~ & SF $\uparrow$ & 9.904 & 11.531 & 13.595 & 12.883 & \textcolor[rgb]{0,0.4470,0.7410}{13.646} & 8.608 & 8.833 & 9.122 & 12.560 & \textcolor{red}{14.047} \\
         ~ & FMI $\uparrow$ & \textcolor{red}{0.868} & 0.858  & 0.858  & 0.851  & 0.848  & 0.839 & 0.847 & 0.850 & \textcolor[rgb]{0,0.4470,0.7410}{0.861} & 0.854 \\
         ~ & VIF $\uparrow$ & 0.229 & 0.458  & \textcolor[rgb]{0,0.4470,0.7410}{0.463} & 0.442  & \textcolor[rgb]{0,0.4470,0.7410}{0.463} & 0.337 & 0.274 & 0.427 & 0.420  & \textcolor{red}{0.465} \\
         ~ & UIQI $\uparrow$ & \textcolor[rgb]{0,0.4470,0.7410}{0.844} & 0.731  & 0.718  & 0.807  & 0.699  & 0.434 & 0.720 & 0.731 & 0.769  & \textcolor{red}{0.860} \\  \hline\hline
        \multirow{6}{*}{LLVIP} & EN $\uparrow$ & 7.126  & 6.812  & 5.296  & 6.924 & 5.263  & 6.861 & 6.365 & 6.708 & \textcolor[rgb]{0,0.4470,0.7410}{7.338} & \textcolor{red}{7.695}  \\
        ~ & AG $\uparrow$ & \textcolor[rgb]{0,0.4470,0.7410}{3.748} & 3.604  & 3.177  & 2.903 & 2.365  & 2.041 & 2.044 & 2.276 & 3.082  & \textcolor{red}{4.676} \\
        ~ & SF $\uparrow$ & \textcolor[rgb]{0,0.4470,0.7410}{13.737} & 13.388 & 12.729 & 9.259 & 10.847 & 5.728 & 7.308 & 7.432 & 10.348 & \textcolor{red}{15.415} \\
        ~ & FMI $\uparrow$ & \textcolor{red}{0.910} & 0.903 & \textcolor[rgb]{0,0.4470,0.7410}{0.904}  & 0.902 & 0.903  & 0.899 & 0.901 & 0.901 & 0.903  & \textcolor[rgb]{0,0.4470,0.7410}{0.904} \\
        ~ & VIF $\uparrow$ & 0.405  & 0.469  & 0.242  & 0.431 & 0.283  & 0.362 & 0.206 & 0.364 & \textcolor[rgb]{0,0.4470,0.7410}{0.470} & \textcolor{red}{1.001} \\
        ~ & UIQI $\uparrow$ & 0.785  & 0.651  & 0.746  & \textcolor{red}{0.848} & 0.275  & 0.436 & 0.769 & \textcolor[rgb]{0,0.4470,0.7410}{0.824} & 0.533  & 0.694  \\  \hline\hline
        \multirow{6}{*}{$M^{3}FD$} & EN $\uparrow$ & \textcolor[rgb]{0,0.4470,0.7410}{7.225} & 6.794  & 6.428  & 6.728  & 6.843  & 6.428 & 6.639 & 6.780 & 7.085  & \textcolor{red}{7.511} \\
        ~ & AG $\uparrow$ & \textcolor[rgb]{0,0.4470,0.7410}{4.247} & 4.218  & 4.239  & 3.640  & 4.240  & 2.411 & 2.769 & 2.719 & 3.671  & \textcolor{red}{5.483} \\
        ~ & SF $\uparrow$ & 12.822 & 12.645 & \textcolor[rgb]{0,0.4470,0.7410}{12.965} & 10.072 & 12.881 & 6.458 & 8.355 & 7.654 & 10.768 & \textcolor{red}{16.150} \\
        ~ & FMI $\uparrow$ & \textcolor{red}{0.891} & 0.887  & \textcolor[rgb]{0,0.4470,0.7410}{0.888} & 0.879  & 0.881  & 0.871 & 0.866 & 0.878 & 0.874  & 0.878  \\
        ~ & VIF $\uparrow$ & 0.349  & 0.495  & 0.329  & 0.445  & \textcolor[rgb]{0,0.4470,0.7410}{0.603} & 0.302 & 0.211 & 0.381 & 0.378  & \textcolor{red}{0.890} \\
        ~ & UIQI $\uparrow$ & 0.721  & 0.714  & \textcolor{red}{0.886} & 0.807  & 0.609  & 0.432 & 0.806 & 0.809 & 0.626  & \textcolor[rgb]{0,0.4470,0.7410}{0.814} \\ 
        \toprule[0.5mm]
    \end{tabular}
    }
\label{Tab.2}
\vspace{-0.4cm}
\end{table*}

\setlength{\parindent}{0.8em}
\textit{2) Quantitative Comparisons:}
We further conduct a quantitative comparison of test images from three datasets to evaluate the effectiveness of GAN-HA. Table \ref{Tab.2} presents the average scores for the six metrics described above, with the best values highlighted in red and the next best in blue. It is evident that GAN-HA achieves the highest scores for AG, SF, and VIF in all scenes, demonstrating its superiority in preserving gradient and texture information, and enhancing visual quality. Moreover, GAN-HA also ranks at the top for EN, FMI, and UIQI metrics in most cases, proving its excellence in information richness, mutual information utilization, and image brightness preservation. While there are some cases where our results are suboptimal, these may be attributed to variations in scene characteristics.

\vspace{-0.2cm}
\subsection{Ablation Experiment}
\label{section:D2}
\vspace{-0.1cm}
In this subsection, a series of experiments are conducted to validate the effectiveness of the GAN-HA structure, which includes two parts: evaluation of the specific structure in the generator, and verification of the rationality of heterogeneous discriminators.

\begin{table}[t]
\setlength{\abovecaptionskip}{0.1cm}
    \centering
    \renewcommand\arraystretch{1}
    \caption{Ablation study of the generator structure on RoadScene. The best results are marked in \textbf{bold}.}
    \resizebox{\linewidth}{!}{
    \begin{tabular}{c|cccccc}
        \toprule[0.5mm]
        \multirow{2}{*}{Methonds} & \multicolumn{6}{c}{RoadScene} \\
        ~ & \cellcolor{black!20}EN$\uparrow$   &                \cellcolor{black!20}AG$\uparrow$   & \cellcolor{black!20}SF$\uparrow$    & \cellcolor{black!20}FMI$\uparrow$  & \cellcolor{black!20}VIF$\uparrow$  & \cellcolor{black!20}UIQI$\uparrow$    \\  \hline\hline        
        w/o Sampling        & 7.206 & 3.513 & 10.139 & \textbf{0.865} & 0.084 & 0.848 \\
        w/o Skip Connection & 7.148 & 4.904 & 13.604 & 0.850 & 0.368 & 0.857 \\
        w/o AFS             & 7.137 & 3.883 & 10.500 & 0.843 & 0.258 & \textbf{0.863} \\ \hline
        GAN-HA              & \textbf{7.293} & \textbf{5.173} & \textbf{14.047} & 0.854 & \textbf{0.465} & 0.860 \\ 
        \toprule[0.5mm]
    \end{tabular}
    }
\label{Tab.3}
\vspace{-0.2cm}
\end{table}

\begin{table}[!t]
\setlength{\abovecaptionskip}{0.1cm}
    \centering
    \renewcommand\arraystretch{1}
    \caption{Ablation study of the generator structure on LLVIP. The best results are marked in \textbf{bold}.}
    \resizebox{\linewidth}{!}{
    \begin{tabular}{c|cccccc}
        \toprule[0.5mm]
        \multirow{2}{*}{Methonds} & \multicolumn{6}{c}{LLVIP} \\
        ~ & \cellcolor{black!20}EN$\uparrow$   &                \cellcolor{black!20}AG$\uparrow$   & \cellcolor{black!20}SF$\uparrow$    & \cellcolor{black!20}FMI$\uparrow$  & \cellcolor{black!20}VIF$\uparrow$  & \cellcolor{black!20}UIQI$\uparrow$    \\  \hline\hline        
        w/o Sampling        & 7.160 & 3.786 & 13.359 & 0.904 & 0.160 & 0.613 \\
        w/o Skip Connection & 7.164 & 3.444 & 11.131 & \textbf{0.905} & 0.561 & 0.665 \\
        w/o AFS             & 6.968 & 2.543 & 8.815  & 0.903 & 0.399 & 0.615 \\ \hline
        GAN-HA              & \textbf{7.695} & \textbf{4.676} & \textbf{15.415} & 0.904 & \textbf{1.001} & \textbf{0.694} \\ 
        \toprule[0.5mm]
    \end{tabular}
    }
\label{Tab.4}
\vspace{-0.2cm}
\end{table}

\begin{table}[!t]
\setlength{\abovecaptionskip}{0.1cm}
    \centering
    \renewcommand\arraystretch{1}
    \caption{Ablation study of the generator structure on $M^{3}FD$. The best results are marked in \textbf{bold}.}
    \resizebox{\linewidth}{!}{
    \begin{tabular}{c|cccccc}
        \toprule[0.5mm]
        \multirow{2}{*}{Methonds} & \multicolumn{6}{c}{$M^{3}FD$} \\
        ~ & \cellcolor{black!20}EN$\uparrow$   &                \cellcolor{black!20}AG$\uparrow$   & \cellcolor{black!20}SF$\uparrow$    & \cellcolor{black!20}FMI$\uparrow$  & \cellcolor{black!20}VIF$\uparrow$  & \cellcolor{black!20}UIQI$\uparrow$    \\  \hline\hline        
        w/o Sampling        & 7.011 & 4.101 & 12.348 & \textbf{0.887} & 0.070 & 0.802 \\
        w/o Skip Connection & 6.932 & 4.137 & 12.168 & 0.878 & 0.507 & \textbf{0.857} \\
        w/o AFS             & 6.638 & 2.690 & 7.904  & 0.864 & 0.263 & 0.854 \\ \hline
        GAN-HA              & \textbf{7.511} & \textbf{5.483} & \textbf{16.150} & 0.878 & \textbf{0.890} & 0.814 \\ 
        \toprule[0.5mm]
    \end{tabular}
    }
\label{Tab.5}
\vspace{-0.4cm}
\end{table}

\setlength{\parindent}{0.8em}
\textit{1) Analysis of the structure of the generator:}
Here the ablation experiments on the multi-scale structure, skip connections, and AFS in the generator are carried out.

\noindent\textbf{Ablation of Sampling.} 
In the generator network, we refer to the multi-scale structure \cite{li2020nestfuse} to extract and fuse the deep features of the image. This is realized by sampling operators. To confirm the effectiveness of this multi-scale structure, we conduct the ablation experiment of sampling operators. In this experiment, other structures and settings remain unchanged, while the sampling operations are eliminated (w/o Sampling), i.e., the generator network becomes a dense skip-connected structure. Table \ref{Tab.3} - Table \ref{Tab.5} provide the specific evaluation of the metrics. It can be seen that, except for the FMI metric of individual scenarios, all other metrics have been improved. The multi-scale structure effectively utilizes deep information of the image. 

\noindent\textbf{Ablation of Skip Connections.} 
In order to reduce information loss during feature extraction and reconstruction, skip connections are also employed in the generator network, so this structure needs to be validated. As indicated in Table \ref{Tab.3} - Table \ref{Tab.5}, the addition of skip connections results in improved performance in EN, AG, SF, and VIF. Thus, the use of skip connections is reasonable. 

\noindent\textbf{Ablation of AFS.} 
To demonstrate the effectiveness of the designed AFS module, the ablation experiment is performed. In Table \ref{Tab.3} - Table \ref{Tab.5}, w/o AFS represents that the proposed fusion strategy is replaced by a simple overlap:

\begin{equation}
F_{fused}^k = 0.5\times F_{ir}^k + 0.5\times F_{vi}^k
\end{equation}
It can be seen that the proposed AFS module notably boosts the metrics including EN, AG, SF, FMI, and VIF, which effectively improves the information expression capability of the fused image.

\begin{figure}[!t]
\setlength{\abovecaptionskip}{0.1cm}
\setlength{\belowcaptionskip}{-0.4cm}
  \centering
  \includegraphics[width=\linewidth]{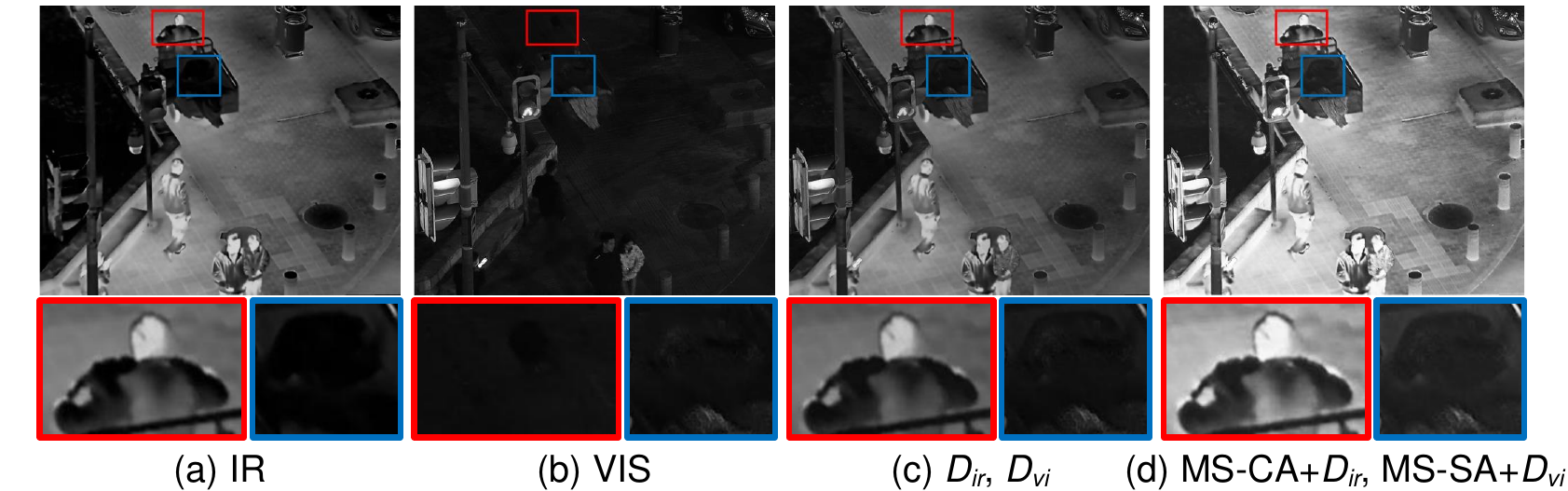}
  \caption{Discussion of the MS-CA and  MS-SA modules on LLVIP.}
  \label{Fig.13}
\end{figure}

\begin{figure}[!t]
\setlength{\abovecaptionskip}{0.1cm}
\setlength{\belowcaptionskip}{-0.4cm}
  \centering
  \includegraphics[width=\linewidth]{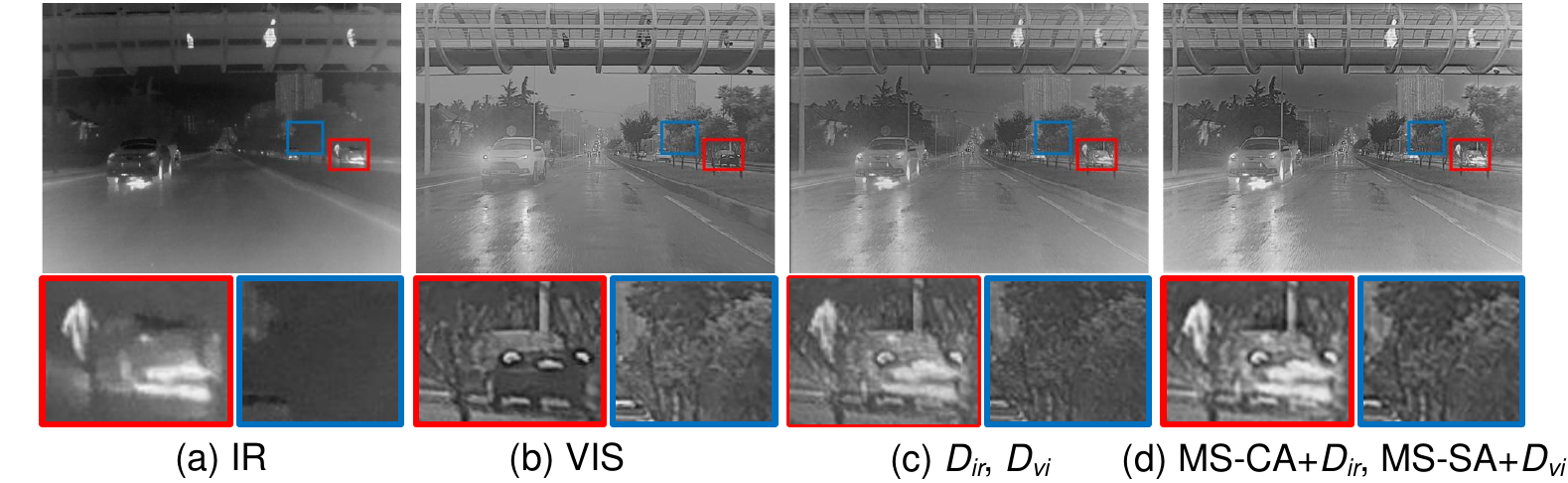}
  \caption{Discussion of the MS-CA and MS-SA modules on $M^{3}FD$.}
  \label{Fig.14}
\end{figure}

\begin{figure}[!t]
\setlength{\abovecaptionskip}{0.1cm}
\setlength{\belowcaptionskip}{-0.4cm}
  \centering
  \includegraphics[width=\linewidth]{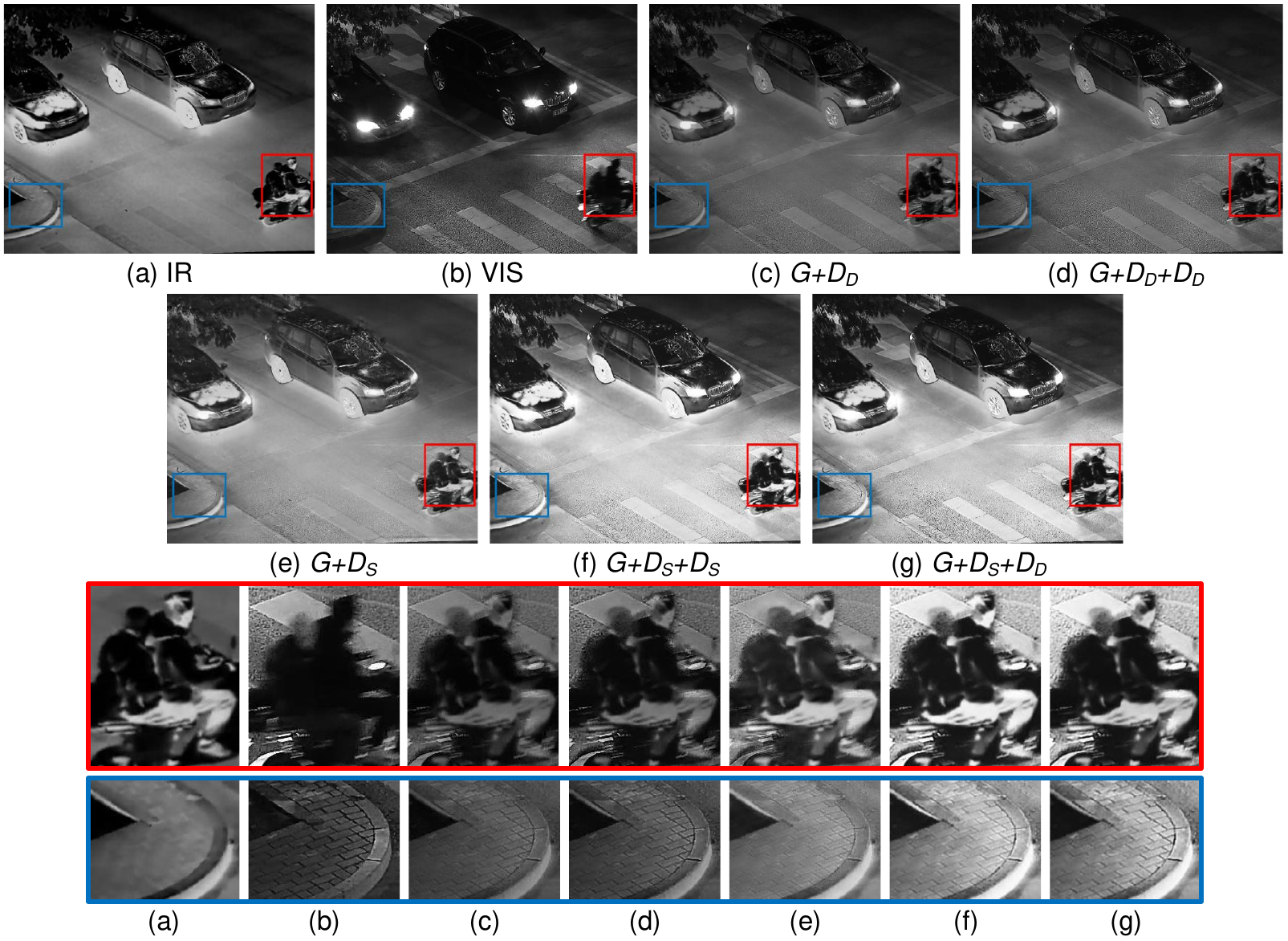}
  \caption{Results for different discriminator structures on LLVIP.}
  \label{Fig.15}
\end{figure}

\begin{figure}[t]
\setlength{\abovecaptionskip}{0.1cm}
\setlength{\belowcaptionskip}{-0.5cm}
  \centering
  \includegraphics[width=\linewidth]{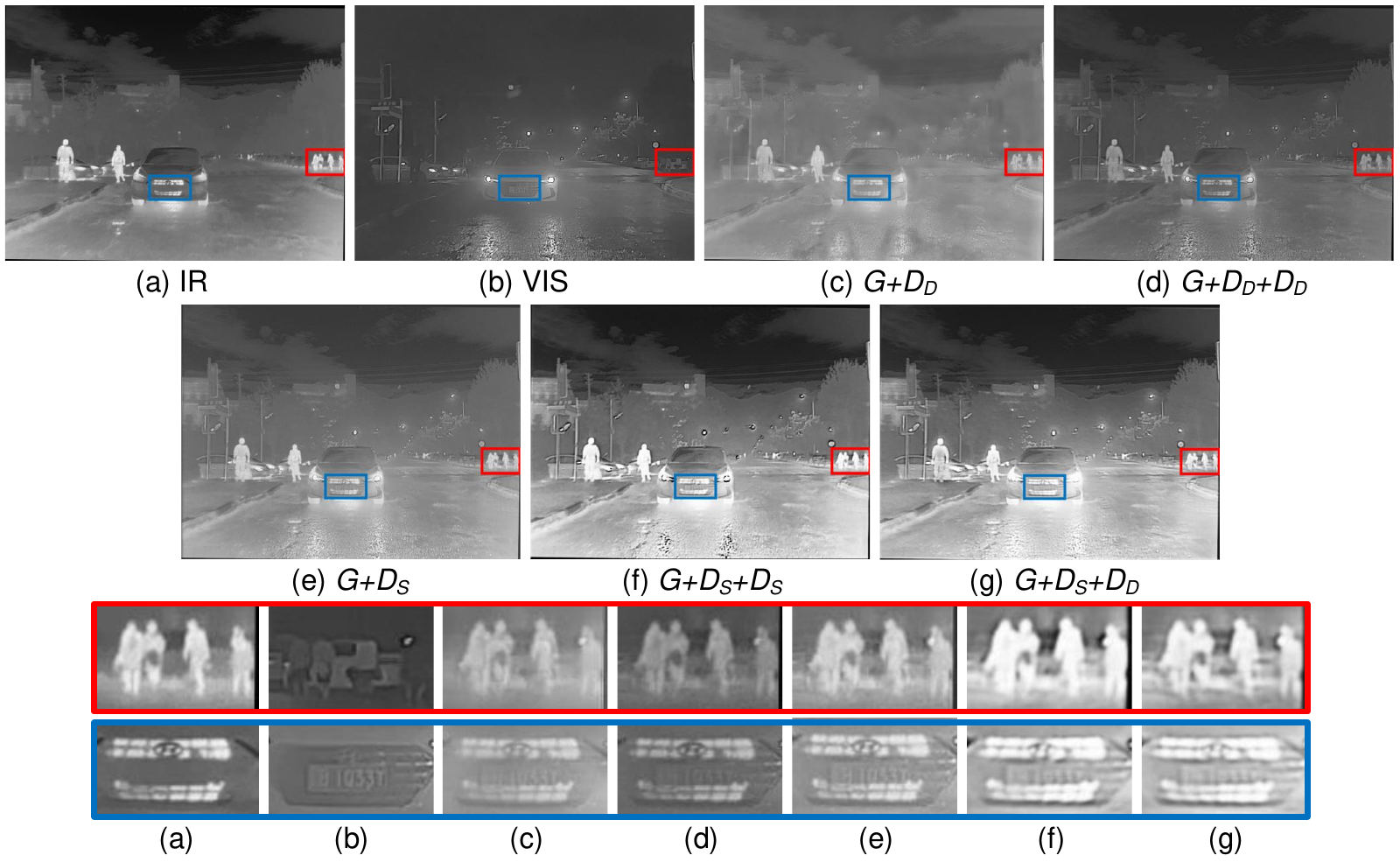}
  \caption{Results for different discriminator structures on $M^{3}FD$.}
  \label{Fig.16}
\end{figure}

\setlength{\parindent}{0.8em}
\textit{2) Analysis of the structure of discriminators:}
In this analysis, we examine the impact of MS-CA and MS-SA, as well as the heterogeneous discriminators \(\mathcal{D}_\mathcal{S}\) and \(\mathcal{D}_\mathcal{D}\).

\noindent\textbf{Ablation of MS-CA and MS-SA.} 
To prompt the model to learn more information about important thermal regions and key local details, we incorporate MS-CA and MS-SA modules. To justify the effectiveness of these modules, an ablation experiment is conducted. In Fig. \ref{Fig.13} and Fig. \ref{Fig.14}, (a) and (b) represent the infrared image and visible image, respectively. (c) is the result after removing the attention modules. (d) presents the proposed setup of GAN-HA. Observing method (c), we find that after discarding the attention modules, the thermal targets are not prominent enough, and the details lack clarity. In contrast, method (d) with the embedded MS-CA and MS-SA modules not only retains rich information learned from the source image but also enhances the contrast and texture details. These results highlight the significance of these two attention modules in discriminators, showcasing their ability to improve the fusion effect further.

\noindent\textbf{Ablation of \(\mathcal{D}_\mathcal{S}\) and \(\mathcal{D}_\mathcal{D}\).} 
GAN-HA includes two heterogeneous discriminators \(\mathcal{D}_\mathcal{S}\) and \(\mathcal{D}_\mathcal{D}\). To demonstrate the effectiveness of each discriminator, we conduct comparison experiments. As depicted in Fig. \ref{Fig.15} and Fig. \ref{Fig.16}, (a) and (b) represent the infrared and visible images, respectively. (c) discards the salient discriminator and only uses \(\mathcal{D}_\mathcal{D}\) and \(G\) for the adversarial game. (d) incorporates detailed discriminators for both infrared and visible images. (e) removes the detailed discriminator and establishes the adversarial game solely between \(\mathcal{D}_\mathcal{S}\) and \(G\). (f) utilizes the same salient discriminator for both infrared and visible images. (g) represents the structure embedded with \(\mathcal{D}_\mathcal{S}\) and \(\mathcal{D}_\mathcal{D}\). To facilitate clear comparisons, we zoom in on the selected thermal region information and texture detail information, respectively, as shown in the bottom red and blue boxes. All parameters and environments remain constant except for the variations mentioned above. Method (c) introduces \(\mathcal{D}_\mathcal{D}\), effectively preserving the textures while the infrared information is lost. Method (d) exhibits richer details after embedding both \(\mathcal{D}_\mathcal{D}\) simultaneously. Method (e) adds \(\mathcal{D}_\mathcal{S}\) on the basis of \(G\), with the distribution of the fused image closer to the infrared image. However, the texture details in the blue boxes are poorly expressed. Method (f) presents more infrared information after introducing two \(\mathcal{D}_\mathcal{S}\), but the detailed information is not clear enough due to excessive brightness. Method (g) shares the same structure as GAN-HA, maintaining high contrast and preserving rich texture details by utilizing both \(\mathcal{D}_\mathcal{S}\) and \(\mathcal{D}_\mathcal{D}\). These comparisons fully justify the use of these two distinct discriminators.

\begin{figure*}[!t]
\setlength{\abovecaptionskip}{-0.1cm}
\setlength{\belowcaptionskip}{-0.6cm}
  \centering
    \subfigure{
    \includegraphics[width=0.2\linewidth]{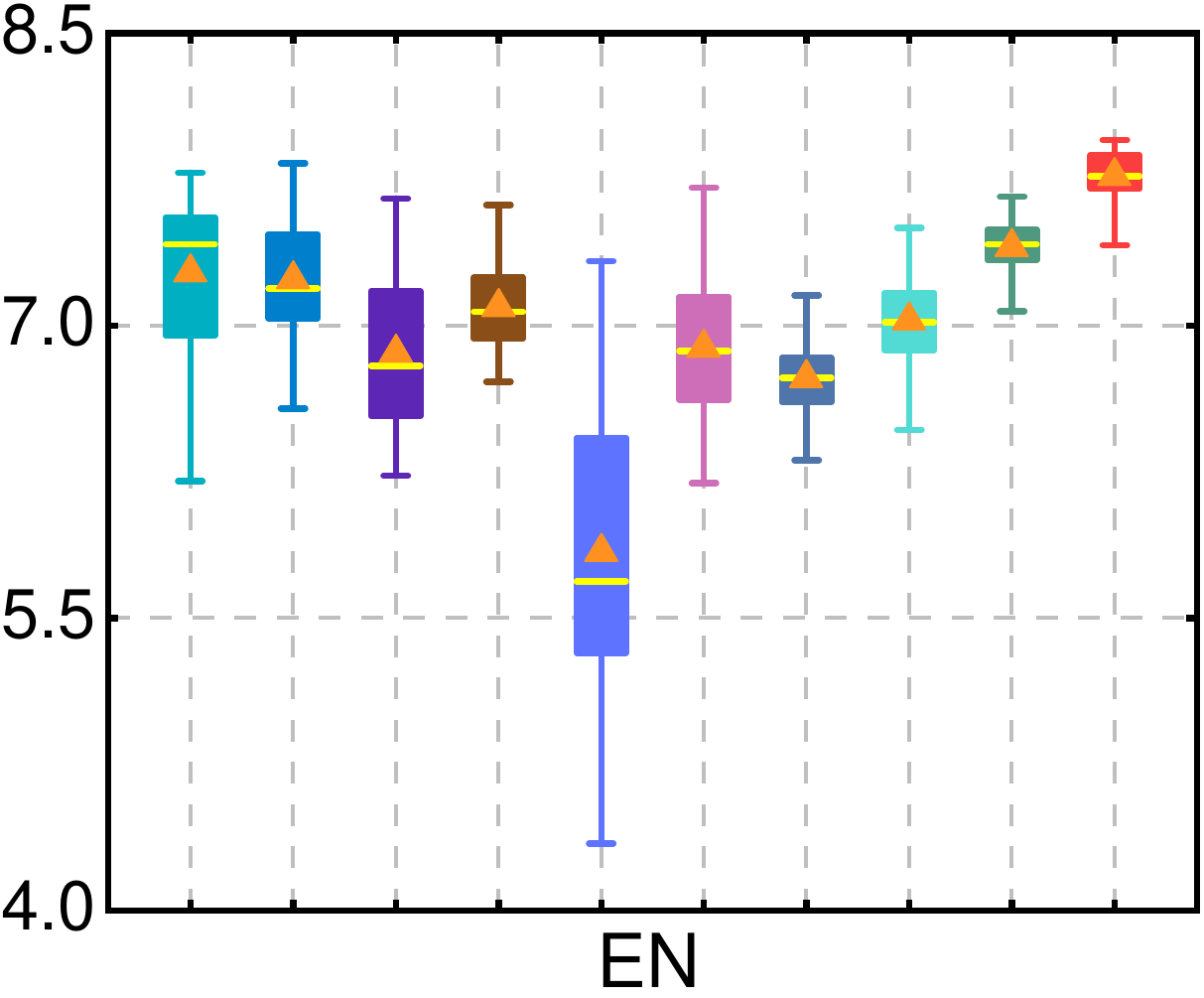}}
    \subfigure{
    \includegraphics[width=0.2\linewidth]{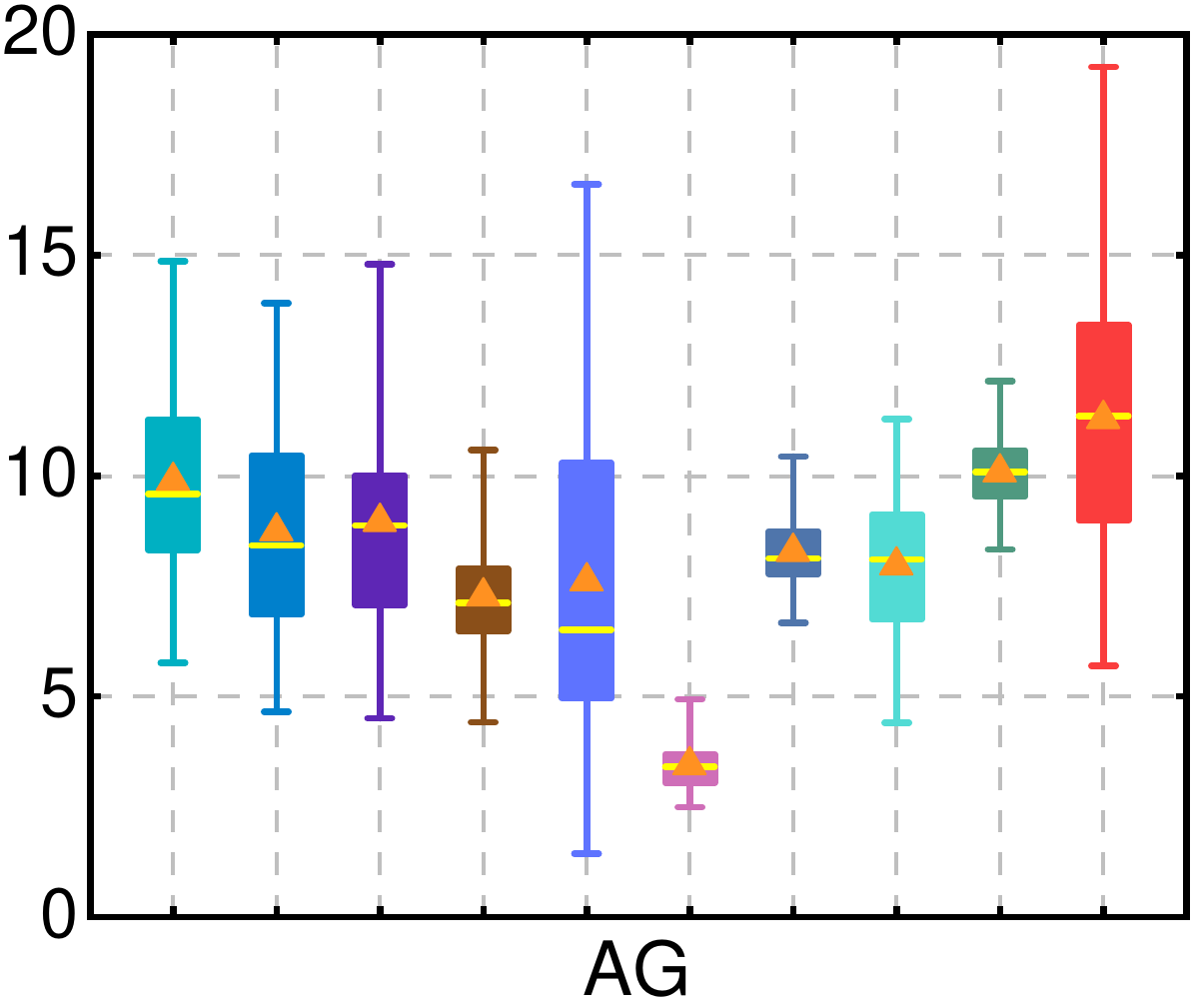}}
    \subfigure{
    \includegraphics[width=0.2\linewidth]{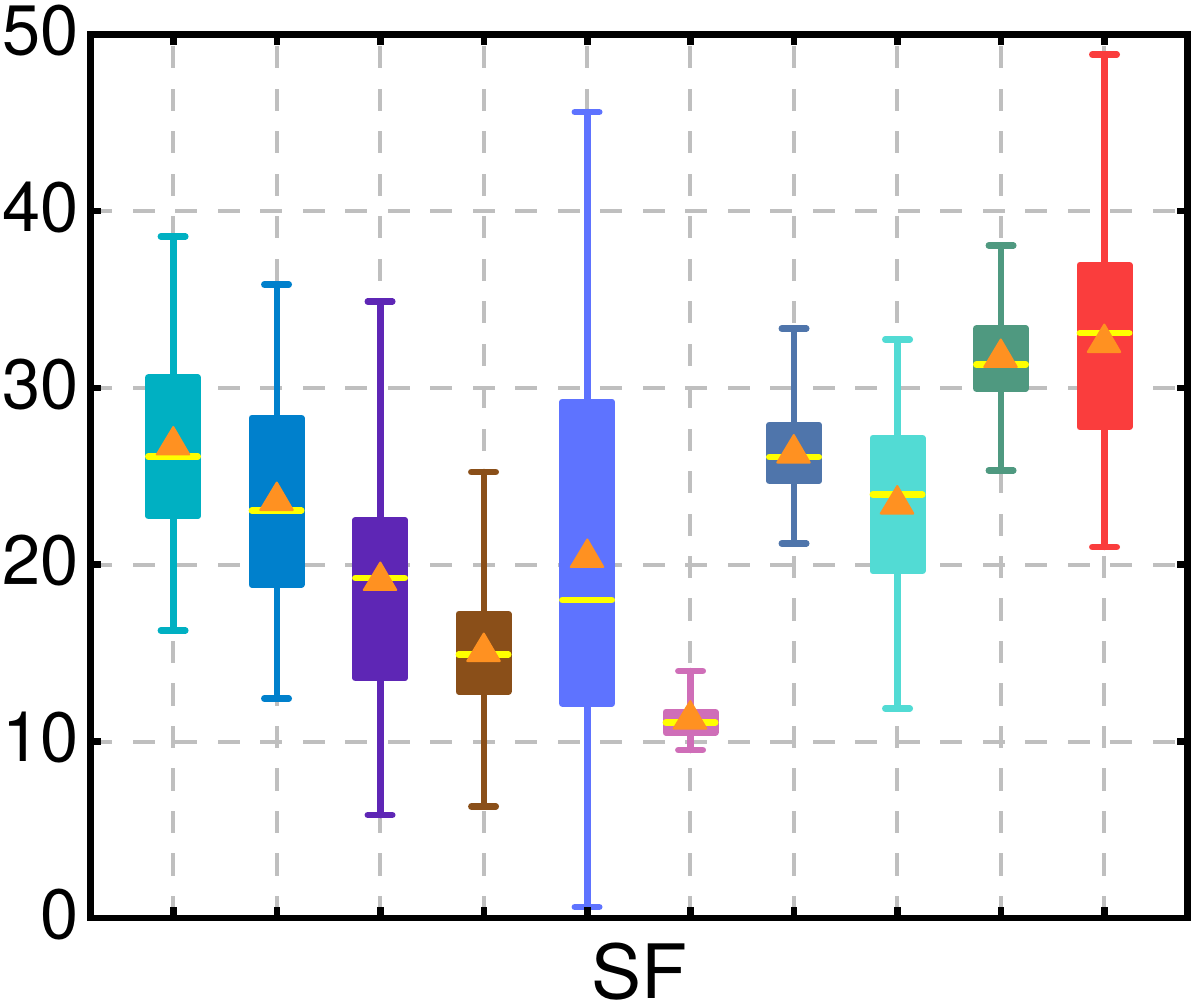}}
    \subfigure{
    \includegraphics[width=0.32\linewidth]{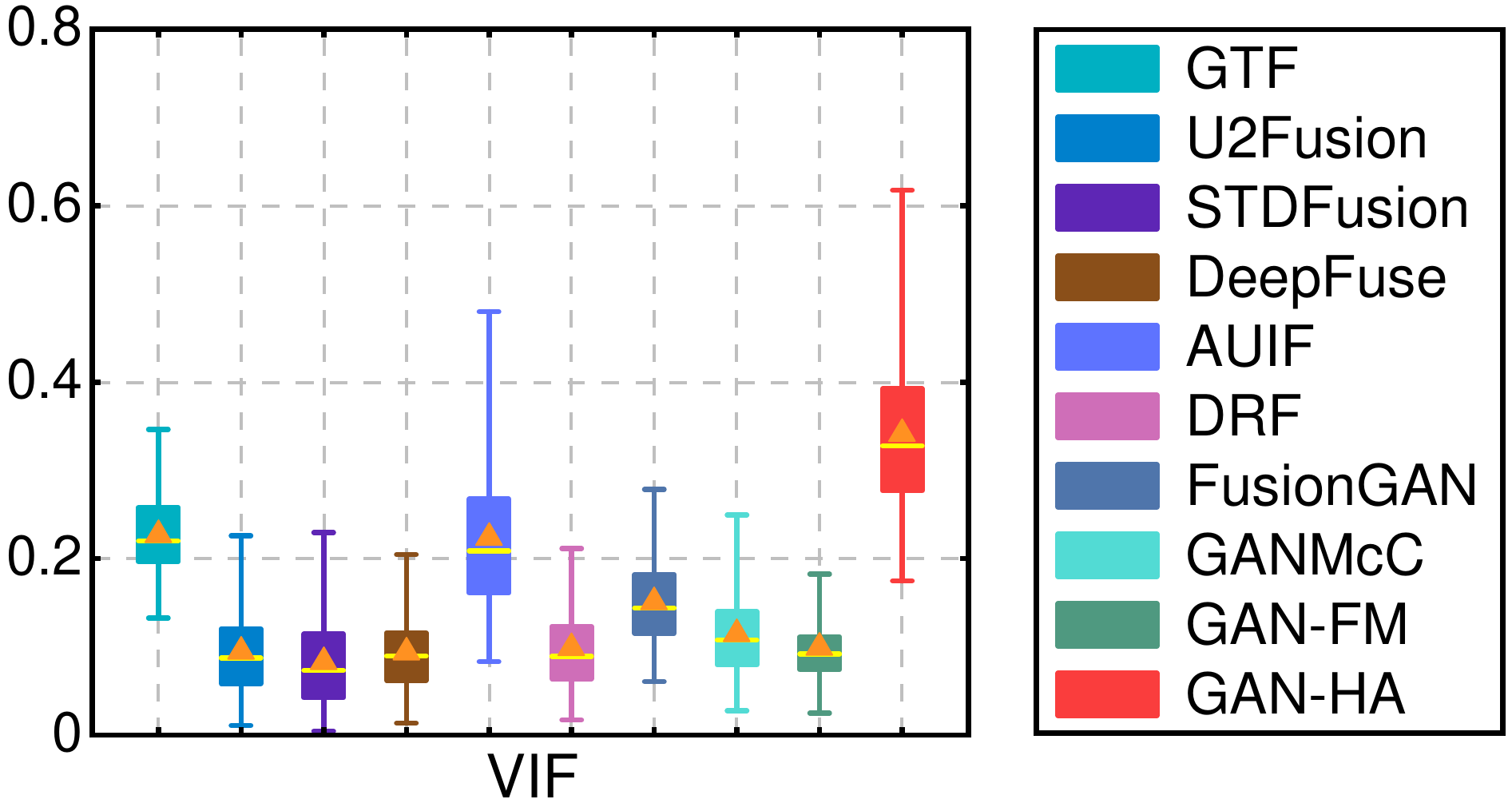}}
  \caption{Degradation analysis under noise scene on LLVIP. From left to right are EN, AG, SF, and VIF. In the boxes, the yellow lines and orange tangles indicate the median and mean values, respectively.}
  \label{Fig.17}
\end{figure*}

\begin{figure*}[!t]
\setlength{\abovecaptionskip}{-0.1cm}
\setlength{\belowcaptionskip}{-0.5cm}
  \centering
    \subfigure{
    \includegraphics[width=0.2\linewidth]{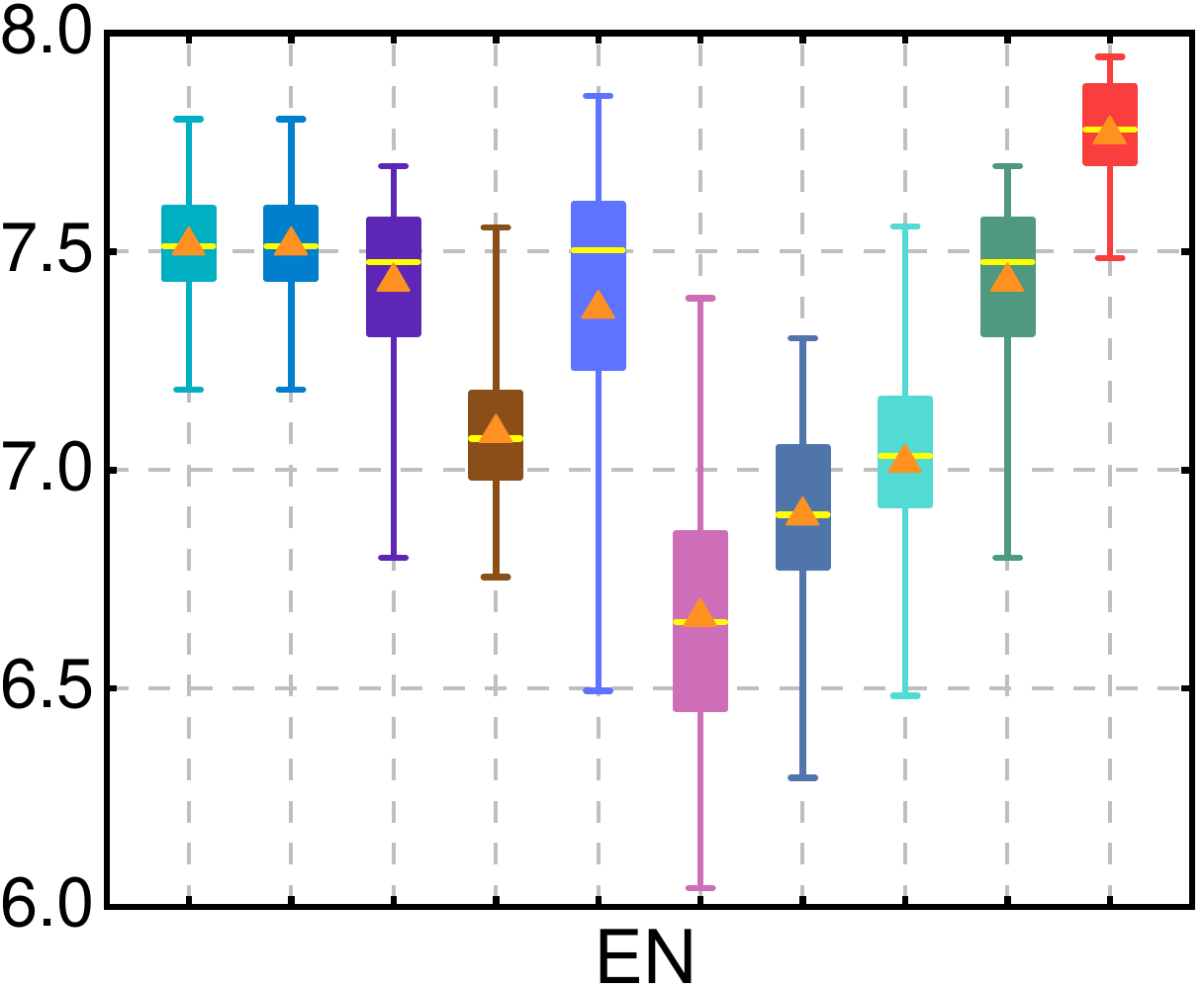}}
    \subfigure{
    \includegraphics[width=0.2\linewidth]{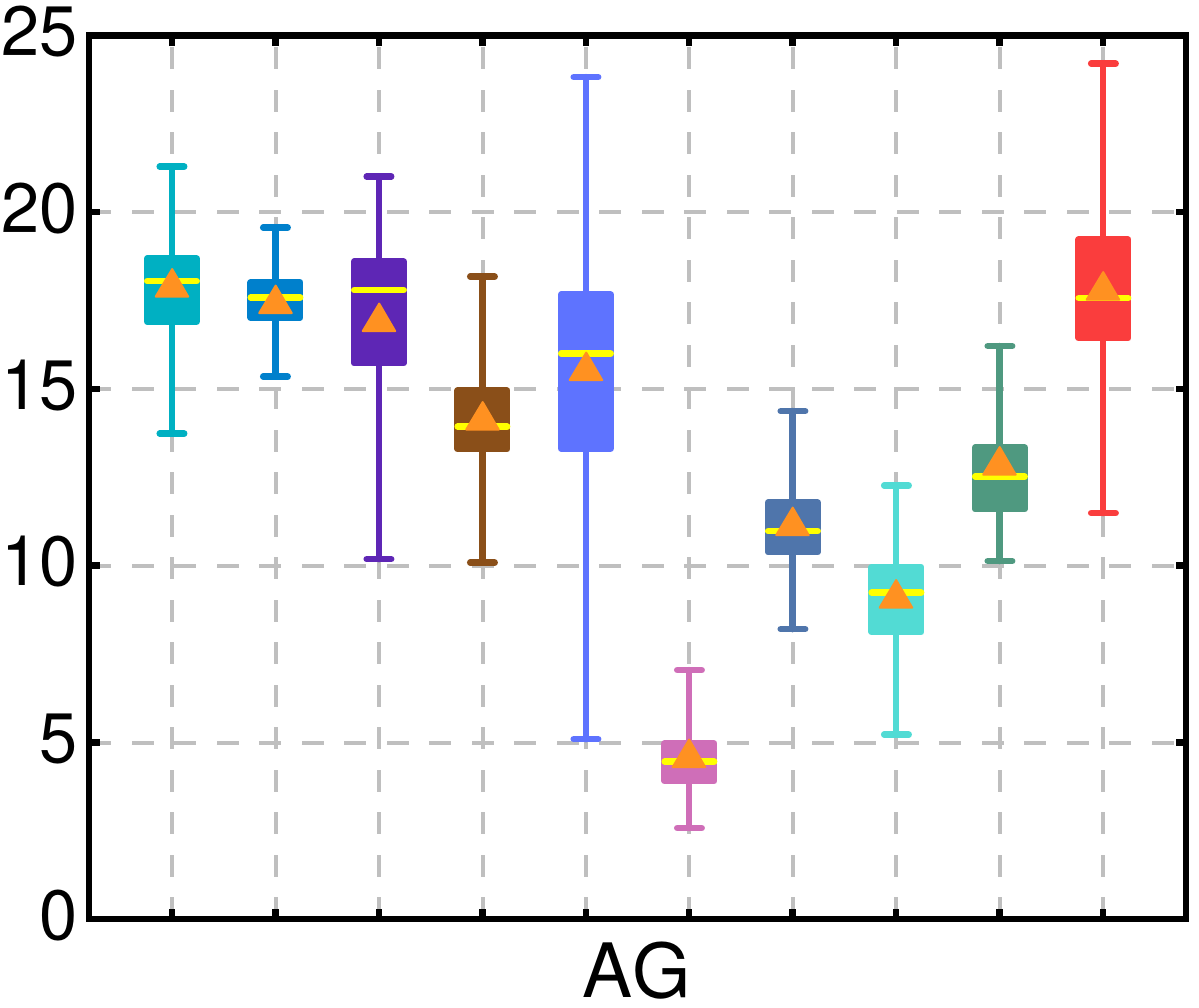}}
    \subfigure{
    \includegraphics[width=0.2\linewidth]{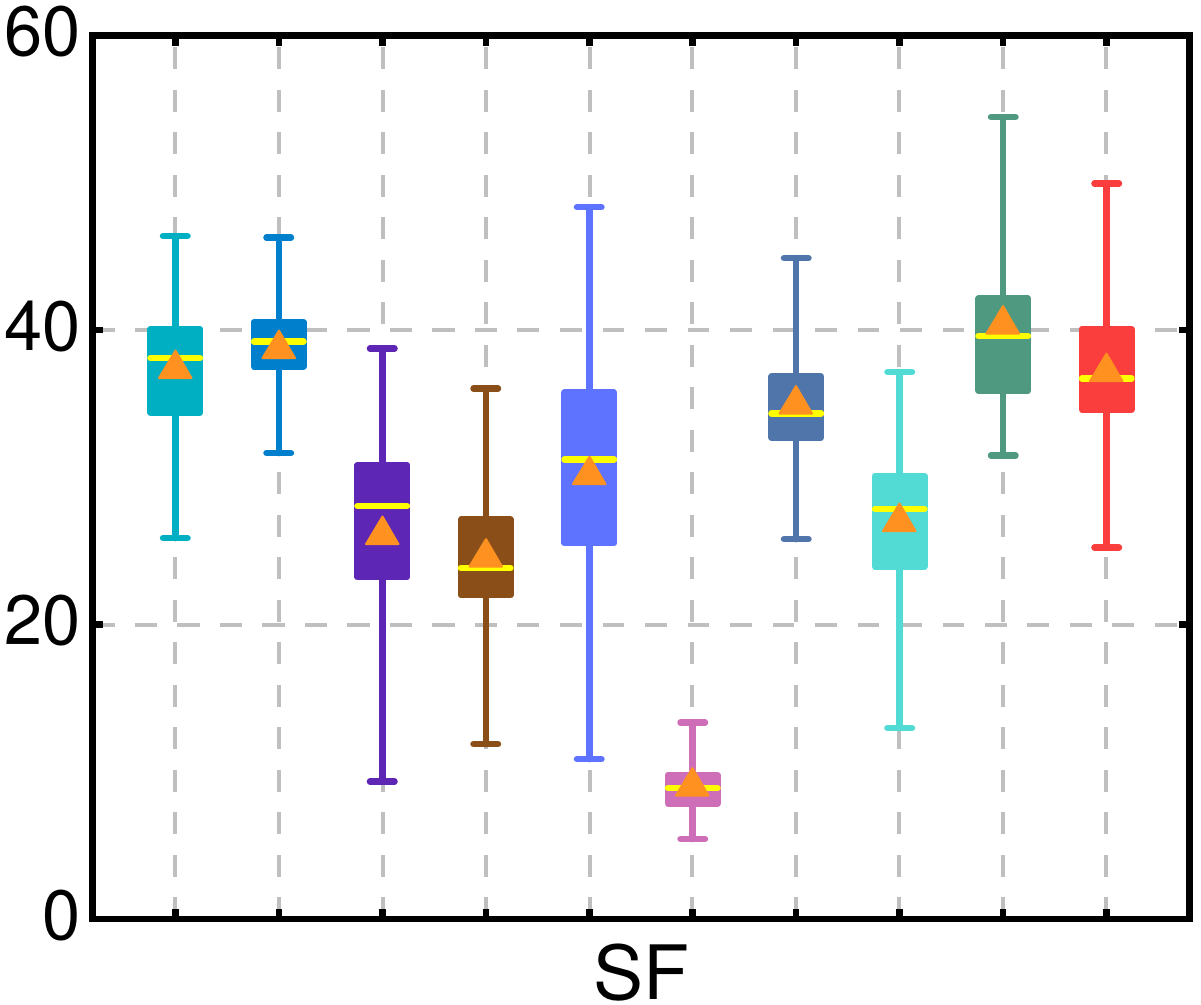}}
    \subfigure{
    \includegraphics[width=0.32\linewidth]{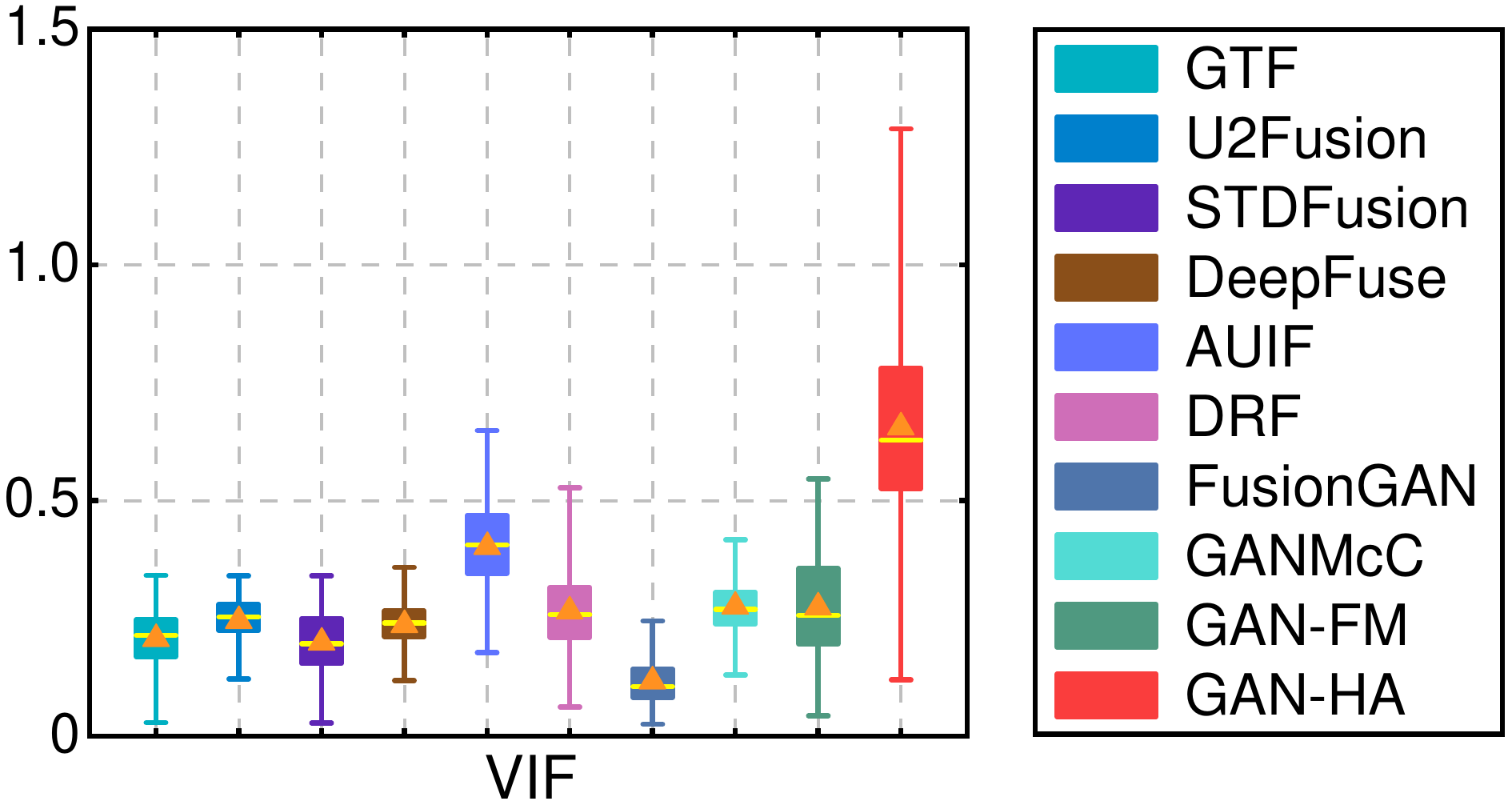}}
  \caption{Degradation analysis under noise scene on $M^{3}FD$. From left to right are EN, AG, SF, and VIF. The yellow lines and orange tangles indicate the median and mean values, respectively.}
  \label{Fig.18}
\end{figure*}

\vspace{-0.2cm}
\subsection{Noise Degradation Resistance Experiment}
\label{section:D3}
\vspace{-0.1cm}
As emphasized in \cite{ma2016infrared}, images captured by visible sensors are susceptible to environmental conditions, which can lead to potential visual degradation—an issue prevalent in practical application scenarios. IVIF technology emerges as a promising solution to mitigate such degradation by integrating information from infrared images. Thus, to delve deeper into the versatility of IVIF for various applications, we investigate the noise degradation resistance capabilities of several SOTA algorithms.

Following the methodology described in \cite{yang2017deep}, we introduce Gaussian noise with a variance \(\sigma\) of 0.03 to the visible images in all testing sets. We evaluate the performance of different methods using four metrics on LLVIP and \(M^3FD\). Fig. \ref{Fig.17} and Fig. \ref{Fig.18} present the noise resistance performance of different algorithms, revealing the overall effectiveness of GAN-HA.

\begin{table*}[!t]
\setlength{\abovecaptionskip}{0.2cm}
    \centering
    \renewcommand\arraystretch{1.2}
    \caption{Comparison of target detection performance on LLVIP and $M^{3}FD$ datasets with yolov5, the best two results are in \textcolor{red}{red} and \textcolor[rgb]{0,0.4470,0.7410}{blue} fonts.}
    \resizebox{\linewidth}{!}{
    \begin{tabular}{p{1.8cm}<{\centering}|p{2.2cm}<{\centering}|p{1.8cm}<{\centering}|p{1.8cm}<{\centering}|p{1.8cm}<{\centering}|p{1.8cm}<{\centering}|p{1.9cm}<{\centering}|p{1.8cm}<{\centering}|p{1.8cm}<{\centering}|p{1.8cm}<{\centering}|p{2.0cm}<{\centering}|p{1.8cm}<{\centering}|p{1.8cm}<{\centering}|p{1.8cm}<{\centering}}
        \toprule[0.5mm]
        Datasets & Metrics & Infrared & Visible & GTF & U2Fusion & STDFuison & DeepFuse & AUIF & DRF & FusionGAN & GANMcC & GAN-FM & GAN-HA \\ \hline\hline
        \multirow{4}{*}{LLVIP} & Precision $\uparrow$ & 0.931 & 0.896 & 0.929 & 0.931 & 0.913 & 0.930 & 0.923 & 0.909 & 0.938 & 0.936 & \textcolor[rgb]{0,0.4470,0.7410}{0.939} & \textcolor{red}{0.955} \\
         ~ & Recall $\uparrow$ & \textcolor{red}{0.886} & 0.808 & 0.849 & 0.880 & 0.839 & 0.878 & 0.865 & 0.805 & 0.842 & 0.876 & \textcolor{red}{0.886} & \textcolor[rgb]{0,0.4470,0.7410}{0.881} \\ 
         ~ & mAP@.5 $\uparrow$ & \textcolor[rgb]{0,0.4470,0.7410}{0.938} & 0.879 & 0.905 & 0.933 & 0.908 & 0.932 & 0.919 & 0.879 & 0.914 & 0.920 & 0.935 & \textcolor{red}{0.943} \\ 
         ~ & mAP@.95 $\uparrow$ & \textcolor[rgb]{0,0.4470,0.7410}{0.543} & 0.433 & 0.472 & 0.521 & 0.487 & 0.508 & 0.505 & 0.462 & 0.496 & 0.487 & 0.517 & \textcolor{red}{0.560} \\  \hline\hline
        \multirow{4}{*}{$M^{3}FD$} & Precision $\uparrow$ & 0.788 & 0.814 & 0.793 & 0.796 & 0.811 & 0.802 & 0.774 & \textcolor{red}{0.860} & 0.775 & 0.778 & 0.742 & \textcolor[rgb]{0,0.4470,0.7410}{0.815} \\
         ~ & Recall $\uparrow$ & 0.525 & 0.525 & 0.547 & 0.585 & 0.543 & 0.570 & 0.575 & 0.500 & 0.531 & 0.557 & \textcolor[rgb]{0,0.4470,0.7410}{0.599} & \textcolor{red}{0.603} \\
         ~ & mAP@.5 $\uparrow$ & 0.567 & 0.597 & 0.603 & \textcolor[rgb]{0,0.4470,0.7410}{0.645} & 0.614 & 0.639 & 0.641 & 0.589 & 0.587 & 0.615 & 0.627 & \textcolor{red}{0.666} \\ 
         ~ & mAP@.95 $\uparrow$ & 0.317 & 0.335 & 0.359 & \textcolor[rgb]{0,0.4470,0.7410}{0.385} & 0.358 & 0.383 & 0.381 & 0.348 & 0.353 & 0.361 & 0.374 & \textcolor{red}{0.406} \\ 
        \toprule[0.5mm]
    \end{tabular}
    }
\label{Tab.6}
\vspace{-0.2cm}
\end{table*}

\vspace{-0.2cm}
\subsection{Downstream Application Verification}
\label{section:D4}
\vspace{-0.1cm}
As mentioned earlier, IVIF combines thermal region information from infrared images with texture information from visible images, which is beneficial for improving the performance of visual tasks. And compared to other SOTA methods, GAN-HA can further emphasize key information. To verify this point, we apply the proposed model to object detection and visual tracking tasks and conduct comparative analyses with other algorithms. 

\setlength{\parindent}{0.8em}
\textit{1) Object Detection:}
Object detection is a widely employed machine vision technique \cite{liu2020deep} for detecting specific objects (people, cars, etc.) in images. Following \cite{liu2022target}, we use the well-known general-purpose detector YOLOv5 for object detection experiments on LLVIP and $M^{3}FD$. Table \ref{Tab.6} reports the results of these two datasets. In this context, Precision denotes the probability of correctly detecting among all detected targets, and higher precision means more real samples are detected. Recall refers to the probability of correctly recognizing all positive samples, and a large recall means fewer targets are lost. The mean average precision (mAP) is a comprehensive consideration of Precision and Recall to evaluate the quality of the model. The value of mAP ranges from 0 to 1, and the closer the value is to 1, the better the model performs. The mAP\textit{@}.5 and mAP\textit{@}.95 denote the mAP values at confidence levels of 0.5 and 0.95, respectively. It is obvious that GAN-HA's fusion results are excellent in these metrics.

\begin{table*}[!t]
\setlength{\abovecaptionskip}{0.1cm}
    \centering
    \renewcommand\arraystretch{1.2}
    \caption{Tracking results for infrared frames, visible frames, and fused frames processed by 10 of the SOTA methods, the best two results are in \textcolor{red}{red} and \textcolor[rgb]{0,0.4470,0.7410}{blue} fonts.}
    \resizebox{\linewidth}{!}{
    \begin{tabular}{p{1.8cm}<{\centering}|p{2.2cm}<{\centering}|p{1.8cm}<{\centering}|p{1.8cm}<{\centering}|p{1.8cm}<{\centering}|p{1.8cm}<{\centering}|p{1.9cm}<{\centering}|p{1.8cm}<{\centering}|p{1.8cm}<{\centering}|p{1.8cm}<{\centering}|p{2.0cm}<{\centering}|p{1.8cm}<{\centering}|p{1.8cm}<{\centering}|p{1.8cm}<{\centering}}
        \toprule[0.5mm]
        Tracker & Metrics & Infrared & Visible & GTF & U2Fusion & STDFuison & DeepFuse & AUIF & DRF & FusionGAN & GANMcC & GAN-FM & GAN-HA \\ \hline\hline
        \multirow{3}{*}{LADCF} & Accuracy $\uparrow$ & 0.5579 & 0.5099 & 0.4720 & 0.4072 & 0.4877 & \textcolor[rgb]{0,0.4470,0.7410}{0.5636} & 0.5428 & 0.4166 & 0.5188 & 0.4668 & 0.4102 & \textcolor{red}{0.5934}  \\
         ~ & Failure $\downarrow$ & 58.9928 & \textcolor{red}{36.7437} & 63.2088 & 68.2383 & 57.7401 & 47.4615 & 47.2419 & 62.7696 & 41.0223 & 48.4960 & 73.4874 & \textcolor[rgb]{0,0.4470,0.7410}{38.4946} \\ 
         ~ & EAO $\uparrow$ & 0.1877 & 0.1970 & \textcolor[rgb]{0,0.4470,0.7410}{0.2094} & 0.1602 & 0.1684 & 0.1848 & 0.1833 & 0.1780 & 0.1929 & 0.1757 & 0.1655 & \textcolor{red}{0.2234} \\ \hline\hline
        \multirow{3}{*}{GFSDCF} & Accuracy $\uparrow$ & 0.5732 & 0.5253 & 0.5853 & 0.4382 & 0.4842 & 0.5751 & 0.5697 & 0.5243 & \textcolor[rgb]{0,0.4470,0.7410}{0.5875} & 0.4900 & 0.4507 & \textcolor{red}{0.6270} \\
         ~ & Failure $\downarrow$ & 54.4705 & \textcolor[rgb]{0,0.4470,0.7410}{31.1229} & 45.7092 & 60.0464 & 52.7589 & 45.4410 & 45.2214 & 55.5957 & 38.1726 & 46.1970 & 60.2955 & \textcolor{red}{30.1482} \\
         ~ & EAO $\uparrow$ & 0.2302 & 0.2498 & 0.2386 & 0.1834 & 0.1918 & 0.2260 & 0.2277 & 0.1988 & \textcolor{red}{0.2606} & 0.1999 & 0.1823 & \textcolor[rgb]{0,0.4470,0.7410}{0.2541} \\
        \toprule[0.5mm]
    \end{tabular}
    }
\label{Tab.7}
\vspace{-0.2cm}
\end{table*}

\setlength{\parindent}{0.8em}
\textit{2) Visual Tracking:}
Visual tracking is also an essential task in computer vision, which aims to localize the position of specific objects in videos. Referring to \cite{li2020mdlatlrr}, we apply the algorithm to the VOT-RGBT sub-challenge, which focuses on short-term tracking in VOT2020 to comparatively evaluate the impact of the fusion approach on tracking performance. The VOT-RGBT benchmark, available in \cite{kristan2019seventh}, comprises 60 video sequences with aligned infrared and visible image pairs. We select the learning adaptive discriminative correlation filter(LADCF) \cite{xu2019learning} and group feature selection and discriminative filter(GFSDCF) \cite{xu2019joint} as trackers for evaluating tracking performance. Among them, LADCF achieves the best performance in the public dataset of the VOT-ST2018 sub-challenge, while GFSDCF further eliminates redundant information based on LADCF through a group feature selection strategy. Their codes are publicly available. For quantitative analysis, we employ three metrics to evaluate tracking performance: Accuracy, Failure, and expected average overlap (EAO). Accuracy represents the average overlap between predicted bounding boxes and ground truth bounding boxes. Failure is a measure of tracker robustness. And EAO evaluates the average overlap in visual properties between the tracker and ground truth. Both Accuracy and EAO are the larger the better, while Failure is the smaller the better. The quantitative analysis is illustrated in Table \ref{Tab.7}. Compared to the metric values of infrared frames and visible frames, our fusion model has a positive impact on RGBT tracking. Furthermore, GAN-HA is more powerful in improving visual tracking performance compared to other fusion methods.

\begin{figure*}[!t]
\setlength{\abovecaptionskip}{0cm}
\setlength{\belowcaptionskip}{-0.4cm}
  \centering
  \includegraphics[width=\linewidth]{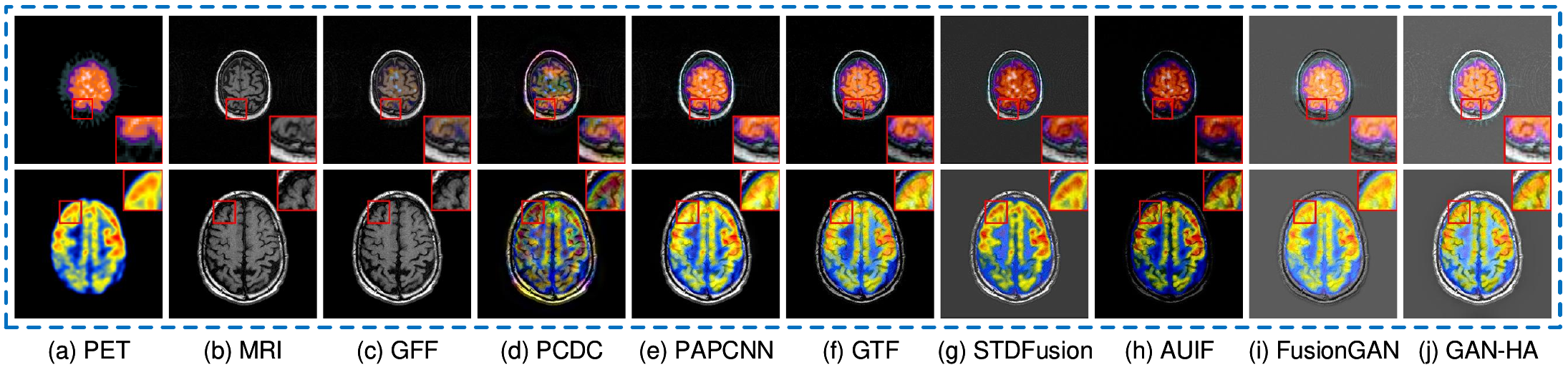}
  \caption{Qualitative results of the medical image fusion task. From left to right, (a) and (b) are PET and MRI images, and (c)-(j) are GFF, PCDC, PAPCNN, GTF, STDFusion, AUIF, FusionGAN, and GAN-HA, respectively.}
  \label{Fig.19}
\end{figure*}

\begin{figure*}[!t]
\setlength{\abovecaptionskip}{0cm}
\setlength{\belowcaptionskip}{-0.5cm}
  \centering
  \includegraphics[width=\linewidth]{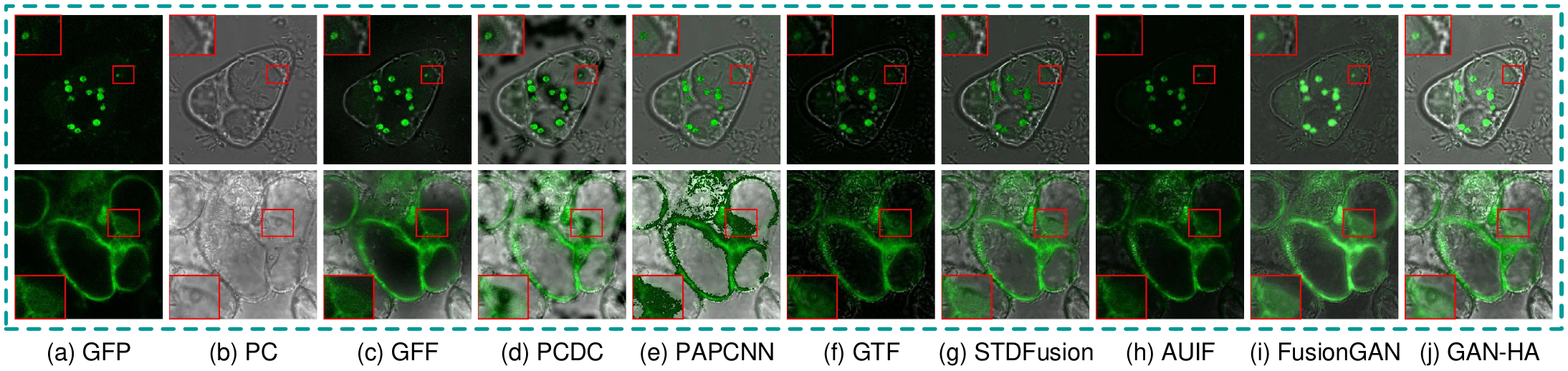}
  \caption{Qualitative results of the biological image fusion task. From left to right, (a) and (b) are GFP and PC images, and (c)-(j) are GFF, PCDC, PAPCNN, GTF, STDFusion, AUIF, FusionGAN, and GAN-HA, respectively.}
  \label{Fig.20}
\end{figure*}

\vspace{-0.2cm}
\subsection{Extended Image Fusion of GAN-HA}
\label{section:D5}
\vspace{-0.1cm}
Since biomedical image fusion shares a similar goal with the IVIF task \cite{james2014medical}, we carry out two comparative experiments to assess the ability of our algorithm for the generalization in the spatial dimension. These tasks are PET and MRI image fusion (PET-MRI) and GFP and PC image fusion (GFP-PC). The images for the first task are sourced from the Whole Brain Atlas (WBA) \cite{vidoni2012whole}, and the second task utilizes data from the ATC dataset \cite{dvovravckova2010attrb1}. Seven algorithms used for comparison are GFF \cite{li2013image}, PCDC \cite{bhatnagar2013directive}, PAPCNN \cite{yin2018medical}, GTF, STDFusion, AUIF, and FusionGAN. Fig. \ref{Fig.19} and Fig. \ref{Fig.20} present the qualitative results of these methods, and it is easy to find that GAN-HA has better fusion outcomes compared to other methods that may suffer from detail distortion and color component preservation challenges. Besides, Table \ref{Tab.8} and Table \ref{Tab.9} also display the average results of these methods on six metrics, confirming that GAN-HA has good generalization performance.

\begin{table}[!t]
\setlength{\abovecaptionskip}{0.1cm}
    \centering
    \renewcommand\arraystretch{1.2}
    \caption{Quantitative analysis results for 259 images of the PET-MRI task, the best two results are in \textcolor{red}{red} and \textcolor[rgb]{0,0.4470,0.7410}{blue} fonts.}
    \resizebox{\linewidth}{!}{
    \begin{tabular}{p{1.8cm}<{\centering}|p{1.8cm}<{\centering}|p{1.8cm}<{\centering}|p{1.8cm}<{\centering}|p{1.8cm}<{\centering}|p{1.9cm}<{\centering}|p{1.8cm}<{\centering}|p{2.0cm}<{\centering}|p{1.8cm}<{\centering}}
        \toprule[0.5mm]
        Metrics & GFF & PCDC & PAPCNN & GTF & STDFusion & AUIF & FusionGAN & GAN-HA \\ \hline\hline
        EN $\uparrow$ & 4.075  & 4.216  & 4.221  & 3.468  & \textcolor[rgb]{0,0.4470,0.7410}{4.254}  & 3.026  & 3.935  & \textcolor{red}{4.805}  \\
        AG $\uparrow$ & 5.952  & 5.929  & \textcolor[rgb]{0,0.4470,0.7410}{5.973}  & 4.914  & 5.391  & 2.954  & 3.232  & \textcolor{red}{5.980}  \\
        SF $\uparrow$ & \textcolor[rgb]{0,0.4470,0.7410}{25.665} & 25.371 & 25.320 & 22.362 & 22.603 & 15.065 & 12.874 & \textcolor{red}{25.680} \\
        FMI $\uparrow$ & \textcolor{red}{0.900}  & \textcolor[rgb]{0,0.4470,0.7410}{0.891}  & 0.889  & 0.877  & 0.875  & 0.854  & 0.852  & 0.876  \\
        VIF $\uparrow$ & 0.493  & 0.367  & \textcolor{red}{0.637}  & 0.446  & 0.259  & 0.365  & 0.265  & \textcolor[rgb]{0,0.4470,0.7410}{0.505}  \\
        UIQI $\uparrow$ & \textcolor[rgb]{0,0.4470,0.7410}{0.907}  & 0.821  & \textcolor{red}{0.922} & 0.753  & 0.498  & 0.534  & 0.454  & 0.639  \\ 
        \toprule[0.5mm]
    \end{tabular}
    }
\label{Tab.8}
\vspace{-0.3cm}
\end{table}

\begin{table}[!t]
\setlength{\abovecaptionskip}{0.1cm}
    \centering
    \renewcommand\arraystretch{1.2}
    \caption{Quantitative analysis results for 147 images of the GFP-PC task, the best two results are in \textcolor{red}{red} and \textcolor[rgb]{0,0.4470,0.7410}{blue} fonts.}
    \resizebox{\linewidth}{!}{
    \begin{tabular}{p{1.8cm}<{\centering}|p{1.8cm}<{\centering}|p{1.8cm}<{\centering}|p{1.8cm}<{\centering}|p{1.8cm}<{\centering}|p{1.9cm}<{\centering}|p{1.8cm}<{\centering}|p{2.0cm}<{\centering}|p{1.8cm}<{\centering}}
        \toprule[0.5mm]
        Metrics & GFF & PCDC & PAPCNN & GTF & STDFusion & AUIF & FusionGAN & GAN-HA \\ \hline\hline
        EN $\uparrow$ & 6.554 & 6.536 & \textcolor[rgb]{0,0.4470,0.7410}{6.628}  & 4.798 & 6.494  & 4.147 & 5.960 & \textcolor{red}{6.914} \\
        AG $\uparrow$ & 3.724 & 3.791 & \textcolor[rgb]{0,0.4470,0.7410}{4.976}  & 2.472 & 4.590  & 1.266 & 2.955 & \textcolor{red}{5.045} \\
        SF $\uparrow$ & 9.580 & 8.498 & \textcolor{red}{16.073} & 6.485 & 11.114 & 4.745 & 7.158 & \textcolor[rgb]{0,0.4470,0.7410}{12.811} \\
        FMI $\uparrow$ & 0.786 & 0.794 & 0.783  & 0.839 & \textcolor{red}{0.862} & \textcolor[rgb]{0,0.4470,0.7410}{0.854} & 0.840 & \textcolor{red}{0.862} \\
        VIF $\uparrow$ & 0.074 & 0.090 & 0.091  & 0.259 & \textcolor[rgb]{0,0.4470,0.7410}{0.486}  & 0.156 & 0.314 & \textcolor{red}{0.760}  \\
        UIQI $\uparrow$ & 0.735 & 0.849 & \textcolor[rgb]{0,0.4470,0.7410}{0.855}  & 0.240 & 0.817  & 0.272 & 0.605 & \textcolor{red}{0.897}  \\ 
        \toprule[0.5mm]
    \end{tabular}
    }
\label{Tab.9}
\vspace{-0.5cm}
\end{table}

\vspace{-0.2cm}
\section{Conclusion}
\label{section:E}
\vspace{-0.1cm}
In this study, we propose a generative adversarial network with a novel heterogeneous dual-discriminator network and a new attention-based fusion strategy. Benefiting from the heterogeneous discriminator structure, our model is capable of learning more features from different source images in a targeted manner, especially the important information. In addition, a new fusion module named attention-based fusion strategy is designed within the generator, which can realize a reasonable representation of the acquired information during the fusion process. Various experiments conducted on numerous public datasets fully demonstrate the superiority of the proposed model, as well as its robustness, potential for practical applications, and generalization. In our future work, we will further explore a lighter and more general fusion model based on the strengths of this model.

\section*{Acknowledgments}
This research was supported by Horizontal Scientific Research Project of Shenzhen Technology University (No. 20231064010094, No. 20231064010029, No. 20241064010018), Graduate School-Enterprise Cooperation Project of Shenzhen Technology University (No. XQHZ202304, No. XQHZ2024006).

\bibliographystyle{elsarticle-num}
\bibliography{ref.bib} 

\end{document}